%% file: bare_jrnl_compsoc.tex
\documentclass[10pt,journal,compsoc]{IEEEtran}

\ifCLASSOPTIONcompsoc
  \usepackage[nocompress]{cite}
\else
  \usepackage{cite}
\fi

\ifCLASSINFOpdf
\else
\fi

\usepackage{ dsfont }
\usepackage{cuted}
\usepackage{capt-of}
\usepackage{verbatim} 

\usepackage{times}
\usepackage{epsfig}
\usepackage{graphicx}
\usepackage{amsmath}
\usepackage{amssymb}
\usepackage{textcomp}
\usepackage{gensymb}
\usepackage{xcolor}
\usepackage{tipa}
\usepackage{hyperref}

\newcommand{\etal}{{et al.~}}
\newcommand{\ie}{{i.e.~}}
\DeclareMathOperator*{\argmin}{argmin}

\newcommand{\cmnt}[1]{\ignorespaces}

\begin{document}
%
\title{Towards a complete 3D morphable model\\ of the human head}
%
%

 \author{Stylianos~Ploumpis,~
         Evangelos~Ververas,~
         Eimear~O'~Sullivan,~
         Stylianos~Moschoglou, 
         Haoyang~Wang,~
         Nick~Pears,~
         William~A.~P.~Smith,~
         Baris Gecer,~
         and~Stefanos~Zafeiriou
 \IEEEcompsocitemizethanks{\IEEEcompsocthanksitem The authors are with the Department
 of Computing, Imperial College London, South Kensington Campus, London SW7 2AZ, UK.
 \IEEEcompsocthanksitem N.~Pears and W.~Smith are with the Department of Computer Science, University of York.
 \protect\\
E-mails: see https://ibug.doc.ic.ac.uk/people, https://cs.york.ac.uk/cvpr/}
 }

%

\markboth{IEEE TRANSACTIONS ON PATTERN ANALYSIS AND MACHINE INTELLIGENCE,~Vol.~X, No.~X, October~2019}%
{Ploumpis \MakeLowercase{\textit{et al.}}: Towards a complete 3D morphable model of the human head}
%

\IEEEtitleabstractindextext{%
\begin{abstract}
Three-dimensional Morphable Models (3DMMs) are powerful statistical tools for representing the 3D shapes and textures of an object class. Here we present the most complete 3DMM of the human head to date that includes face, cranium, ears, eyes, teeth and tongue. To achieve this, we propose two methods for combining existing 3DMMs of different overlapping head parts: i.~use a regressor to complete missing parts of one model using the other, ii.~use the Gaussian Process framework to blend covariance matrices from multiple models. Thus we build a new combined face-and-head shape model that blends the variability and facial detail of an existing face model (the LSFM) with the full head modelling capability of an existing head model (the LYHM). Then we construct and fuse a highly-detailed ear model to extend the variation of the ear shape. Eye and eye region models are incorporated into the head model, along with basic models of the teeth, tongue and inner mouth cavity. The new model achieves state-of-the-art performance. We use our model to reconstruct full head representations from single, unconstrained images allowing us to parameterize craniofacial shape and texture, along with the ear shape, eye gaze and eye color.

\end{abstract}

\begin{IEEEkeywords}
3DMM, Morphable Model combination, 3D reconstruction, craniofacial 3DMM.
\end{IEEEkeywords}}

\maketitle

\IEEEdisplaynontitleabstractindextext
%
\IEEEpeerreviewmaketitle
\IEEEraisesectionheading{\section{Introduction}\label{sec:introduction}}


\begin{figure*}
    \centering
    \includegraphics[width=1\textwidth]{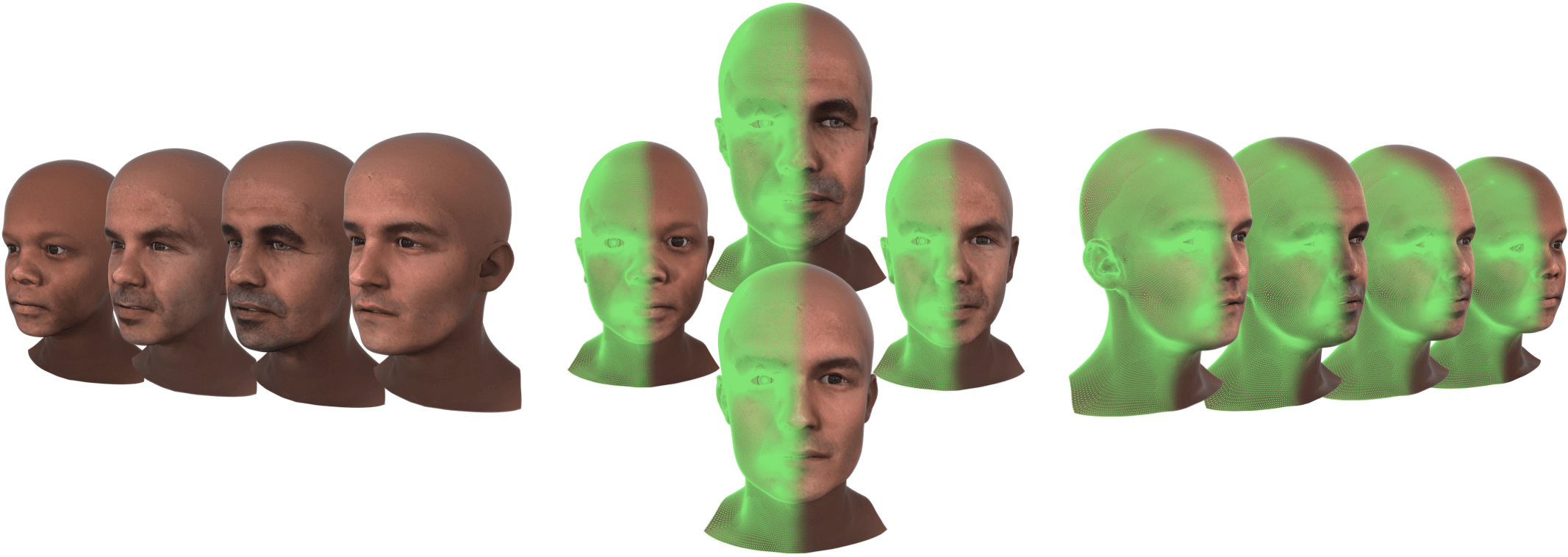}
    \caption{A collection of arbitrary complete head reconstructions from unconstrained single images. Our work aims to combine the most important attributes of the human head (\ie face, cranium, ears, eyes), in order to synthesize novel and realistic 3D head models from data deficient sources.}
    \label{fig:intro_fig}
\end{figure*}

\IEEEPARstart{D}{ue} to their ability to infer and represent 3D surfaces, 3D Morphable Models (3DMMs) have many applications in computer vision, computer graphics, biometrics, and medical imaging \cite{blanz2003face,hu2016face,aldrian2013inverse,staal2015describing}. Many registered raw 3D images (`scans') are required for correctly training a 3DMM, which comes at a very large cost of manual labour for collecting and annotating such images with meta data. Sometimes, only the resulting 3DMMs become available to the research community, and not the raw 3D images. This is particularly true of 3D images of the human face/head, due to increasingly stringent data protection regulations. Furthermore, even if 3DMMs have overlapping parts, their resolution and ability to express detailed shape variation may be quite different, and we may wish to capture the best properties of multiple 3DMMs within a single model. However, it is currently extremely difficult to combine and enrich existing 3DMMs with different attributes that describe distinct parts of an object without such raw data. Therefore, we present a general approach that can be employed to combine 3DMMs from different parts of an object class into a single 3DMM. Due to their widespread use in the computer vision community, we fuse 3DMMs of the human face and the full human head as our exemplar. We add detailed models of the ears, eyes and eye regions to our head model, along with a basic model of the oral cavity, tongue and teeth. Thus we create a \emph{large-scale, full-head} morphable model that has a more complete representation of shape variation than any other published to date. The technique is readily extensible to incorporate detailed models of the human body \cite{hasler2009statistical,allen2003space}, and indeed is applicable to any object class well-described by 3DMMs. Recent works that aim at predicting the 3D representation of more than one morphable model \cite{romero2017embodied, joo2018total}, try to solve this problem with a part-based approach where multiple separate models are fitted and then linearly blended into the final result. Our framework aims at avoiding any discontinuities that might appear from part-based approaches by fusing all models into one single morphable model.
 
More specifically, although there have been many models of the human face both in terms of identity \cite{huber2016multiresolution,zhu2015discriminative,zhu2016face} and expression \cite{bronstein2003expression,zhu2015high}, very few deal with the complete head anatomy \cite{dai20173d}. Building a high-quality, large-scale statistical model that describes the anatomy of the full human head paves directions across numerous disciplines. First, it will assist craniofacial clinicians in diagnosis, surgical planning, and assessment. Second, generating proportionally correct head models based on the geometry of the face will aid computer graphics designers to create realistic avatar-like representations. Third, ergonomic design of headwear, eyewear, breathing apparatus and so on benefits from accurate models of craniofacial shape variation across the population. Finally, a head model will give opportunities that aim at reconstructing a full head representation from data-deficient sources, such as 2D images.

Our key contributions are: (i) a methodology that aims to fuse shape-based 3DMMs, using the human face, head and ear as an exemplar. In particular, we propose both a regression method based on latent shape parameters, and a covariance combination approach, utilized in a Gaussian process framework, (ii) a combined large-scale statistical model of the human head in terms of ethnicity, age and gender that is significantly more accurate than any other existing head morphable model - we make this publicly-available \footnote{Project url: \url{https://github.com/steliosploumpis/Universal_Head_3DMM}} for the benefit of the research community, including versions with and without eyes and teeth, and (iii) an application experiment in which we utilize the combined 3DMM to perform full head reconstruction from unconstrained single images. 


The remainder of the paper is structured as follows. In Section \ref{sec:relatedwork} we review relevant related work.
In Section \ref{sec:face_and_head} we elaborate on the methodology of the face and head model combination and in Sections \ref{sec:ear_modeling}, \ref{sec:eye_modeling} we describe the modeling of ears and eyes, which results in our complete head representation. In Section \ref{sec:texture_modeling_completion},  we describe our head texture completion pipeline and in Section \ref{sec:experiments}, we outline a series of quantitative and qualitative experiments. Finally, conclusions are presented in Section \ref{sec:conclusions}.
\section{Related work}\label{sec:relatedwork}

A very recent survey \cite{egger20193d} identified more complete statistical modelling of the human head as an important open challenge in the development of 3DMMs. Motivated by this goal, we begin by surveying existing attempts to model the face, the full craniofacial region, eyes and ears.  An earlier version of the work in this paper was originally presented in \cite{ploumpis2019combining}. Here, we have extended the model by additionally integrating detailed ear and eye models and a full head texture model as well as including further experimental evaluation.

\subsection{Face models}
\label{sec:faces}
The first 3DMM was proposed by Blanz and Vetter \cite{blanz1999morphable}. They were the first to to recognize the generative capabilities of a 3DMM and they proposed a
technique to capture the variations of 3D faces. Only 200 scans were used to build the model (100 male and 100 female) where dense correspondences were computed based on optical flow that depends on an energy function that describes both the shape and texture. The Basel Face Model (BFM) is the most widely-used and well-known 3DMM, which was built by Paysan \etal \cite{paysan20093d} and utilizes a better registration method than the original Blanz-Vetter 3DMM. They use a known template mesh in which all the vertices have known positions and then they register it to the training scans by utilizing an optimal step Non-rigid Iterative Closest Point algorithm (NICP) \cite{amberg2007optimal}. Standard PCA was employed as a dimensionality reduction technique to construct their model.

Recently, Booth \etal \cite{booth20163d} built a Large-scale Face Model (LSFM) by utilizing nearly $10,000$ face scans. The model is constructed by applying a weighted version of the optimal-step NICP algorithm \cite{de2010optimal}, followed by a Generalized Procrustes Analysis (GPA) and standard PCA. Due to the large number of facial scans, a robust automated procedure was carried out including 3D landmark localization and error pruning of badly registered scans. This work was the first to introduce bespoke models in terms of age, gender and ethnicity, and is the most information-rich 3DMM of face shapes in neutral expression produced to date.

Applications such as 3D model-based reconstruction \cite{booth20173d, gecer2019ganfit} and expression estimation \cite{chang2019deep,chang2018expnet} in 2D images have greatly encouraged the advancement of statistical face models. With the advent of deep neural networks, several recent approaches aimed to extend the traditional 3DMM by replacing the linear models with non-linear approaches \cite{tran2019towards, tran2018nonlinear} or by estimating the coefficients of a 3DMM from a single image \cite{tuan2017regressing}. The main scope of this work lies in combining 3DMMs and building a united representation of the most significant parts of the human head (\ie face, cranium, ears and eyes), rather than creating alternatives for non-linear 3D face reconstruction.

\subsection{Head models}
\label{sec:heads}
In terms of 3DMMs associated with the human body, the main focus of the research literature has been on the reconstruction of the human face, but not other parts of the human head. The reason for this is mainly due to the lack of 3D image datasets that describe the other parts of the human head.
 In recent years, a few works such as \cite{li2017learning} have tried to tackle this task, in which a total of $3,800$ head scans was utilized from the US and European CEASAR body scan database \cite{robinette2002civilian} to build a statistical model of the entire head. The aim of this work focuses mainly on the temporal registration of 3D scans rather than on the topology of the head area. The data consists of full body scans and the resolution in which the head topology was recorded in is insufficient to depict correctly the shape of each individual human head. In addition, the template used for registration in this method is extremely sparse with only $5,000$ vertices which makes it difficult to accurately represent the entire head. Moreover, the registration process incorporates coupling weights for the back of head and the back of the neck, which drastically constrains the actual statistical variation of the entire head area. An extension of this work is proposed in \cite{ranjan2018generating} in which a non-linear model is constructed using convolution mesh autoencoders focusing on facial expressions, but still it lacks the statistical variation of the full cranium.
  Similarly, in the work of Hu and Saito \cite{hu2017avatar}, a full head model is created from single images mainly for real-time rendering. The work aims at creating a realistic avatar model which includes 3D hair estimation. The head topology is considered to be unchanged for all subjects and only the face part of the head is a statistically-correct representation.
 
The most accurate craniofacial 3DMM of the human head both in terms of shape and texture, is the Liverpool-York Head model (LYHM) \cite{dai20173d}. In this work, global craniofacial 3DMMs and demographic sub-population 3DMMs were built from 1,212 distinct identities. They have proposed a dense correspondence system, combining a hierarchical parts-based template morphing framework in the shape channel and a refining optical flow in the texture channel. 
Although this work is the first that describes the statistical correlation between the cranium and the face part, it lacks detail of the facial characteristics, as the spatial resolution of the facial region is not significantly higher than the cranial region. In effect, the variance of the cranial and neck areas dominates that of the facial region in the PCA parameterization. Also, although the model describes how the cranium is affected given the age of the subject, it is biased in terms of ethnicity, due to the lack of ethnic diversity in the dataset.
\subsection{Eye and ear models}
\label{sec:ears_eyes_literature}
There are some key structures of the human head that have an important contribution to the appearance and identity of a person, that perhaps should be treated with greater attention and detail to such an extent that separate 3DMMs should be formulated.

One of the most significant structures of the human head are the eyes, by which we communicate and and through their movements we expresses our interests, our attention, and our emotional disposition. As a result, eye appearance\cite{hansen2009eye} and gaze estimation \cite{zhang2015appearance,park2018deep} are active topics in computer vision. The first parametric approach to eye modeling was proposed by B\'erard \etal \cite{berard2016lightweight} where a 3DMM model was build by utilizing a database of eyeball scans \cite{berard2014high}. Although the results of the reconstruction were appealing in terms of quality, the method for reconstruction appeared to be semi-automatic. The most recent 3DMM of the human eye was proposed in \cite{wood20163d} focusing on the eyeball as well as on the peripheral eye region and skin region around the eye. In our work, instead of treating the eye region as a separate model we globally estimate the position of the eyes and, by employing sparse localized deformation blendshapes, we are able to determine the gaze direction and the general shape of the eye region.

Another structure of the human head that contributes to biometric recognition and to the general appearance of a person are the ears \cite{pflug2012ear, abaza2013survey}. Numerous works have been published over the years on ear-based recognition \cite{islam2011efficient,yuan2010ear}, thus making the ear an important structure to represent in any human head modeling. The two foremost examples of 3DMMs of the ear are those of Zolfghari \etal \cite{zolfghari2016ear} and Dai \etal \cite{dai2018ear}. Both models were constructed by applying PCA to ear meshes from the SYMARE database \cite{jin2014symare}, using $58$ and $20$ samples respectively. To overcome the limited statistical variation of their restricted sample size, \cite{dai2018ear} estimate the 3D shape of ears in a landmarked 2D ear image dataset and combine these with their initial model to propose a data-augmented 3DMM. Both the LSFM face model and the LYHM head model templates contain the ear; however, modelling the \emph{detailed} shape of the ear was not the intention during the construction of either of these. As such, the statistical variation of the ear is limited in both models, and neither contain a sufficient number of vertices in the ear region to accurately represent its complex structure. In this work, we enrich the statistical variability of the aforementioned models by fusing our own ear model constructed from high-resolution ear scans. To the best of our knowledge, the resulting head model is the most complete and accurate 3DMM of the human head.

\begin{figure*}
    \centering
    \includegraphics[width=1\textwidth]{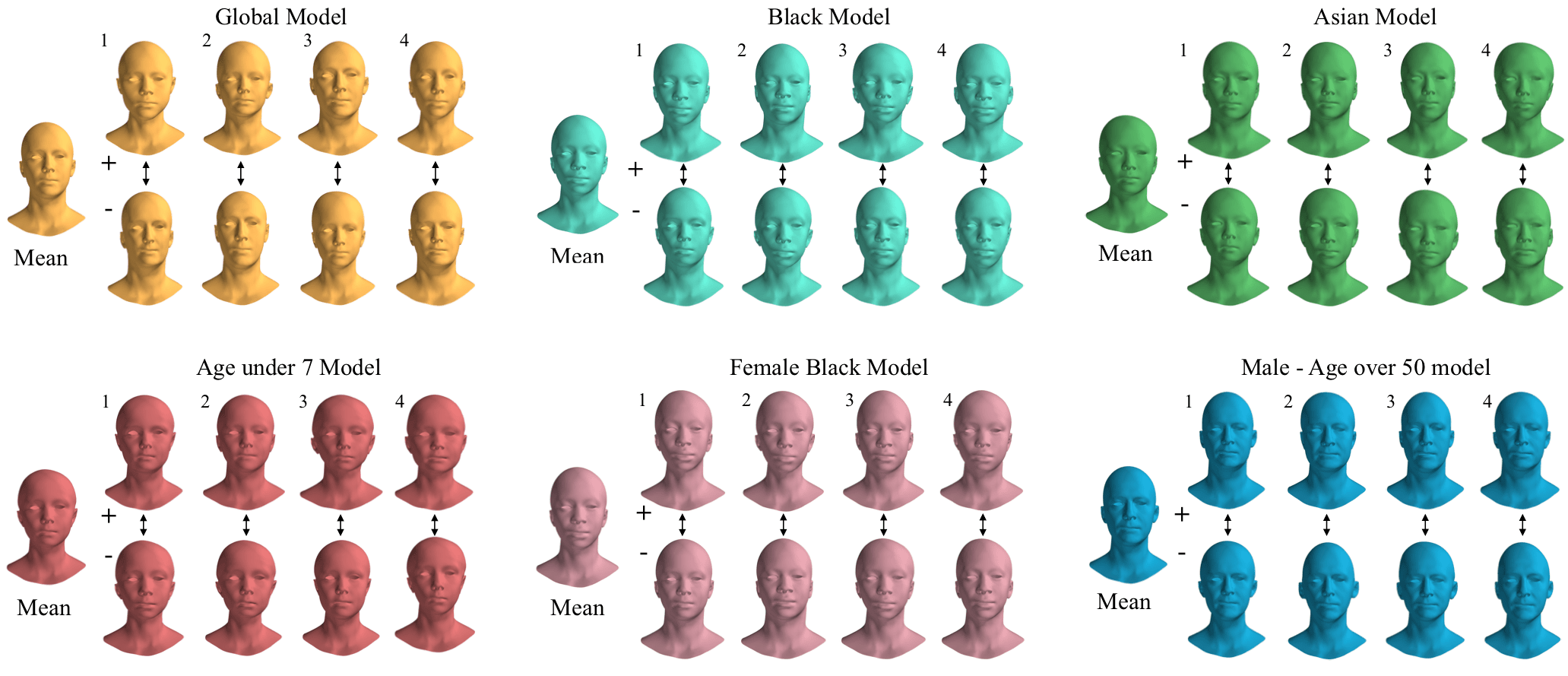}
    \caption{The bespoke Combined Face \& Head Models. Visualisation of the first four shape components along with the mean head shapes. Due to the large demographic information of LSFM we are able to construct bespoke combined head model for any given age, gender or ethnicity group.}
    \label{fig:CHFM_componets}
\end{figure*}

\section{Face and head shape combination}
\label{sec:face_and_head}
In this section, we propose two methods to combine the LSFM face model with the LYHM full head model. The first approach, utilizes the latent PCA parameters and solves a linear least squares problem to approximate the full head shape, whereas the second constructs a combined covariance matrix that is later utilized as a kernel in a Gaussian Process Morphable Model (GPMM) \cite{luthi2017gaussian}.

\subsection{Regression modelling}
\label{reg_combination}
\begin{figure*}[h]
\begin{center}
\includegraphics[width=1\linewidth]{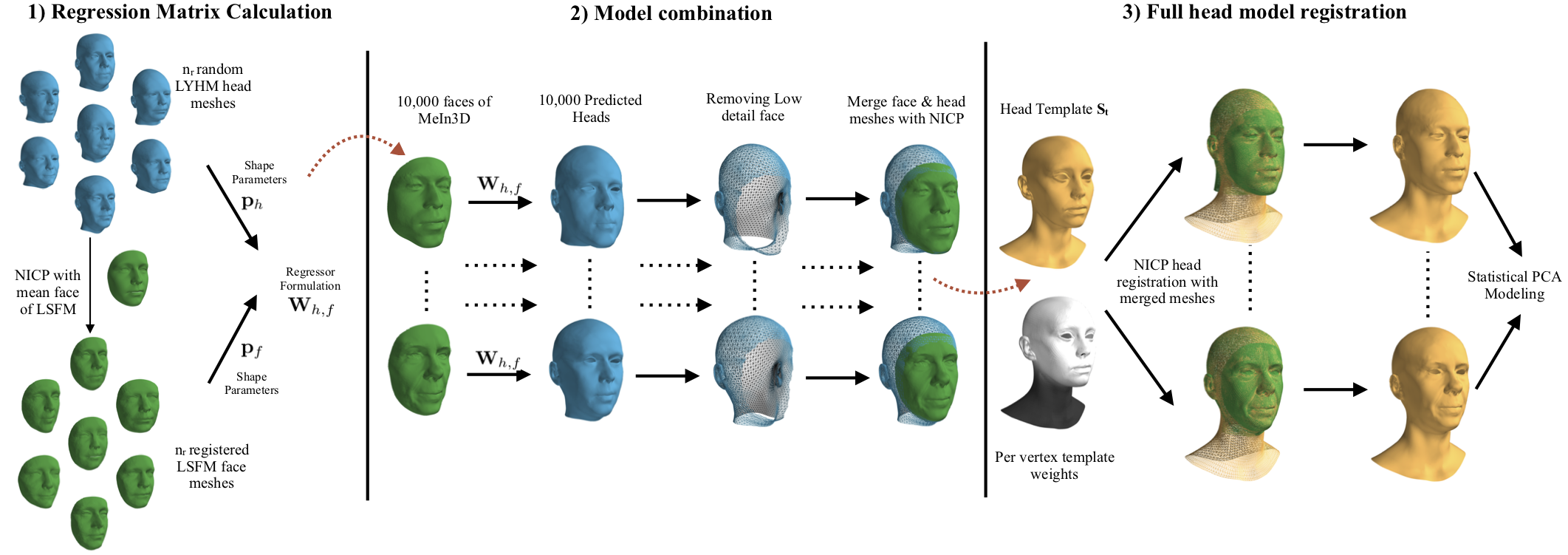}
\caption{The regression modeling pipeline. 1) The left part illustrates the  matrix formulation from the original LYHM head model; 2) the central part demonstrates how we utilize the \emph{MeIn3D} database to produce highly-detailed head shapes; 3) the final part on the right depicts the registration framework along with the per-vertex template weights and the statistical modeling.}
\label{fig:reg_diagram}
\end{center}
\end{figure*}

Figure~\ref{fig:reg_diagram} illustrates the three-stage regression modeling pipeline, which comprises 1) regression matrix calculation, 2) model combination and 3) full head model registration followed by PCA modeling. Each stage is now described.

For stage 1, let us denote the 3D mesh (shape) of an object with $N$ points as a $3N \times 1$ vector
\begin{equation}
    \mathbf{S} = [\mathbf{x}_1^T\dots\mathbf{x}_N^T]^T = [x_1, y_1, z_1,\dots x_N, y_N, z_N]^T
\end{equation}
The LYHM is a PCA generative head model with $N_h$ points, described by an orthonormal basis after keeping the first $n_h$ principal components $\mathbf{U}_{h} \in \mathds{R}^{3N_h\times n_h} $ and the associated $\mathbf{\lambda}_{h}$ eigenvalues. This model can be used to generate novel 3D head instances as follows:
\begin{equation}
\mathcal{\mathbf{S}}_{h}(\mathbf{p}_{h}) = {\mathbf{m}}_{h} + {\mathbf{U}}_{h} \mathbf{p}_{h}
\label{equ:head_shape_instance}
\end{equation}
where $ \mathbf{p}_{h} = \left[p_{h_1} \ldots p_{h_{n_h}} \right ]^T$ are the $n_h$ shape parameters. Similarly the LSFM face model with $N_f$ number of points, is described by a corresponding orthonormal basis after keeping the $n_f$ principal components $\mathbf{U}_f \in \mathds{R}^{3N_f\times n_f} $ and the associated $\mathbf{\lambda}_f$ eigenvalues. The model generates novel 3D faces instances by:
\begin{equation}
\mathcal{\mathbf{S}}_{f}(\mathbf{p}_{f}) = {\mathbf{m}}_{f} + {\mathbf{U}}_{f} \mathbf{p}_{f}
\label{equ:face_shape_instance}
\end{equation}
where $ \mathbf{p}_{f} = \left[p_{f_1} \ldots p_{f_{n_f}} \right ]^T$ are the $n_f$ shape parameters. 

In order to combine the two models, we synthesize data directly from the latent eigenspace of the head model ($\mathbf{U}_{h}$) by drawing random samples from a Gaussian distribution defined by the principal eigenvalues of the head model. The standard deviation for each of the distributions is equal to the square root of the eigenvalue. In that way we produce randomly $n_r$ distinct shape parameters.


After generating the random full head instances we apply non-rigid registration (NICP) \cite{de2010optimal} between the head meshes and the cropped mean face of the LSFM face model. We perform this task in each one of the $n_r$ meshes in order to get the facial part of the full head instance and describe it in terms of the LSFM topology. Once we acquire those registered meshes we project them to the LSFM subspace and we retrieve the corresponding shape parameters. Thus, for each one of the randomly produced head instances, we have a pair of shape parameters ($\mathbf{p}_{h}, \mathbf{p}_{f}$) corresponding to the full head representation and to the facial area respectively.

By utilizing those pairs we construct a matrix $\mathbf{C}_h \in\mathds{R}^{n_h\times n_r }$ where we stack all the head shape parameters and a matrix $\mathbf{C}_f \in\mathds{R}^{n_f\times n_r }$ where we stack the face shape parameters from the LSFM model. We would like to find a matrix $\mathbf{W}_{h,f} \in \mathds{R}^{n_h\times n_f } $ to describe the mapping from the LSFM face shape parameters $\mathbf{p}_{f}$ to the corresponding LYHM full head shape parameters $\mathbf{p}_{h}$. We solve this by formulating a linear least square problem that minimizes:
\begin{equation}
 \left\lVert \mathbf{C}_h - \mathbf{W}_{h,f} \mathbf{C}_f \right\rVert^2
\label{equ:normal_equ_head_face}
\end{equation}
By utilizing the normal equation, the solution of \eqref{equ:normal_equ_head_face} is readily given by:
\begin{equation}
\mathbf{W}_{h,f} = \mathbf{C}_h\mathbf{C}_f^T \left( \mathbf{C}_f\mathbf{C}_f^T \right)^{-1}
\end{equation}
where $\mathbf{C}_f^T \left( \mathbf{C}_f\mathbf{C}_f^T \right)^{-1}$ is the right pseudo-inverse of $\mathbf{C}_f$. Given a 3D face instance $\mathbf{S}_f$, we derive the 3D shape of the full head, $\mathbf{S}_h$, as follows:
\begin{equation}
\mathcal{\mathbf{S}}_h ={\mathbf{m}}_{h} + \mathbf{U}_{h} \mathbf{W}_{h,f} \mathbf{U}_{f}^T\left( \mathbf{S}_f - {\mathbf{m}}_{f}\right)
\end{equation}
In this way we can map and predict the shape of the cranium region for any given face shape in terms of LYHM topology. 

In stage 2 (Figure~\ref{fig:reg_diagram}), we employ the large MeIn3D database \cite{booth20163d} which includes nearly $10,000$ 3D face images, and we utilize the $\mathbf{W}_{h,f}$ regression matrix to construct new full head shapes that we later combine with the real facial scans. We achieve this by discarding the facial region of the the full head instance which has less detailed information and we replace it with the registered LSFM face of the MeIn3D scan. In order to create a unique instance we merge the meshes together by applying a NICP framework, where we deform only the outer parts of the facial mesh to match with the cranium angle and shape so that the result is a smooth combination of the two meshes. Following the formulation in \cite{de2010optimal}, this is accomplished by introducing higher stiffness weights in the inner mesh (lower on the outside) while we apply the NICP algorithm. To compute those weights we measure the Euclidean distance of a given point from the nose tip of the mesh and we assign a relative weight to that point. The bigger the distance from the nose tip, the smaller the weight of the point.

One of the drawbacks of the LYHM is the arbitrary neck circumference, where the neck tends to get broader when the general shape of the head increases. In stage 3 (Figure~\ref{fig:reg_diagram}), we aim at excluding this factor from our final head model by applying a final NICP step between the merged meshes and our head template $\mathbf{S}_t$. We utilized the same framework as before with the point-weighted strategy where we assign weights to the points based on their Euclidean distance from the center of the head mass. This helps us avoid any inconsistencies of the neck area that might appear from the regression scheme. For the area around the ear, we have introduced 50 additional landmarks to control the registration and preserve the general shape of the ear area.

After applying our pipeline to each one of the $10,000$ meshes, we perform PCA on the points of the mesh and we acquire a new generative full head model that exhibits more detail in the face area in combination with bespoke head shapes.

\subsection{Gaussian process modeling}
\label{sec:GP_modeling}
Gaussian processes for model combination is a less complicated and more robust technique that does not generate irregular head shapes due to poor regression values.

The concept of Gaussian Process Morphable Models (GPMMs) was recently introduced in \cite{luthi2017gaussian,gerig2018morphable,koppen2018gaussian}. The main contribution of GPMMs is the generalization of classic Point Distribution Models (such as are constructed using PCA), with the help of Gaussian processes. A shape is modeled as a deformation $u$ from the reference shape $\mathbf{S}_R$ i.e. a shape can be represented as: 
\begin{equation}
    \mathbf{S} = \lbrace \mathbf{x} + u(\mathbf{x}) | \mathbf{x}\in \mathbf{S}_R \rbrace
\end{equation} 
where $u$ is a deformation function $u : \Omega \rightarrow \mathds{R}^{3}$ with $\Omega \supseteq  \mathbf{S}_R$. The deformations are modeled as a Gaussian process $u \sim \mathcal{GP}\left(\mu, k\right)$. Where $\mu : \Omega \rightarrow \mathds{R}^{3} $ is the mean deformation and $k : \Omega \times \Omega  \rightarrow \mathds{R}^{3\times3}$ is a covariance function or kernel. 

The Gaussian process model is capable of operation outside of the space of valid face shapes. This depends highly on the kernels chosen for this task. In the classic approaches, the deformation function is learned through a series of typical example surfaces $\mathbf{S}_1, \ldots,\mathbf{S}_n$ where a set of deformation fields is learned $\lbrace u_1,\ldots, u_n\rbrace, u_i(\mathbf{x})  : \Omega \rightarrow \mathds{R}^{d}$ where $u_i(\mathbf{x})$ denotes the deformation field that maps a point $\mathbf{x}$ on the reference shape to the corresponding point on the $i_{th}$-training surface. 

A Gaussian process $\mathcal{GP}\left( \mu_{PDM},k_{PDM}\right)$ that models this characteristic deformations is obtained by estimating the empirical mean: 
\begin{equation}
    \mu_{PDM}\left(\mathbf{x}\right) = \frac{1}{n}\sum_{i=1}^n u_{i}(\mathbf{x})
\end{equation}
and the covariance function:
\begin{equation}
\begin{split}
    k_{PDM}\left( \mathbf{x}, \mathbf{y}\right) = \frac{1}{1-n}\sum_{i=1}^n \left( u_{i}(\mathbf{x}) - \mu_{PDM}(\mathbf{x}) \right) \\
    \left( u_{i}(\mathbf{y}) - \mu_{PDM}(\mathbf{y}) \right)^T
\end{split}
\end{equation}
This kernel is defined as the empirical/sample covariance kernel. This specific Gaussian process model is a continuous analog to a PCA model and it operates in the facial deformation spectrum. In our case we are lacking with regards to the original head scans so we are unable to learn deformation fields from them, nor combine them with the MeIn3D facial dataset. In order to overcome this problem, we have utilized the already-learned point distribution models. 
For each one of the models (LYHM, LSFM), we know the principal orthonormal basis and the eigenvalues. Hence the covariance matrix for each model is defined:
\begin{equation} \label{equ:compute_covariance}
\begin{split}
    \mathbf{K}_h = \mathbf{U}_h \mathbf{ \Lambda}_h  \mathbf{U}_h^T \\
    \mathbf{K}_f = \mathbf{U}_f \mathbf{ \Lambda}_f  \mathbf{U}_f^T 
\end{split}
\end{equation}
where $\mathbf{K}_h\in \mathds{R}^{3N_h\times 3N_h}$ and $\mathbf{K}_f\in \mathds{R}^{3N_f\times 3N_f}$ are the covariance matrices, and the $\mathbf{\Lambda}_h\in \mathds{R}^{n_h\times n_h}$ and $\mathbf{\Lambda}_f\in \mathds{R}^{n_f\times n_f}$ are diagonal matrices with the eigenvalues in their the main diagonal of the head and face model respectively.

We aim at constructing a universal covariance matrix $\mathbf{K}_{U} \in \mathds{R}^{3N_U\times 3N_U}$ that accommodates the high detailed facial properties of the LSFM and the head distribution from the LYHM. We keep, as a reference, the mean of the head model and we non-rigidly register the mean face of the LSFM. Both PCA models must be in the same scale space for this method to work, which was not necessary for the regression method. Similarly, we register our head template $\mathbf{S}_t$ by utilizing the same pipeline as before for full head registration, which is going to be used as the reference mesh for the new joined covariance matrix.


For each point pair $i,j$ in $\mathbf{S}_t$, there exists a local covariance matrix $\mathbf{K}_U^{i,j} \in \mathds{R}^{3\times3}$. In order to calculate its value, we begin by projecting the points onto the mean head mesh. If both points lie outside the face area that the registered mean mesh of LSFM covers, we identify their exact location in the mean head mesh in terms of barycentric coordinates $(c_1^i, c_2^i, c_3^i)$ for the $i_{th}$ point and $(c_1^j, c_2^j, c_3^j)$ for the $j_{th}$ point with respect to their corresponding triangles $\mathbf{t}_i = [\mathbf{v}_1^T,\mathbf{v}_2^T,\mathbf{v}_3^T]^T,\mathbf{t}_j = [\mathbf{k}_1^T,\mathbf{k}_2^T,\mathbf{k}_3^T]^T$.

\begin{figure}[h]
\begin{center}
\includegraphics[width=1\linewidth]{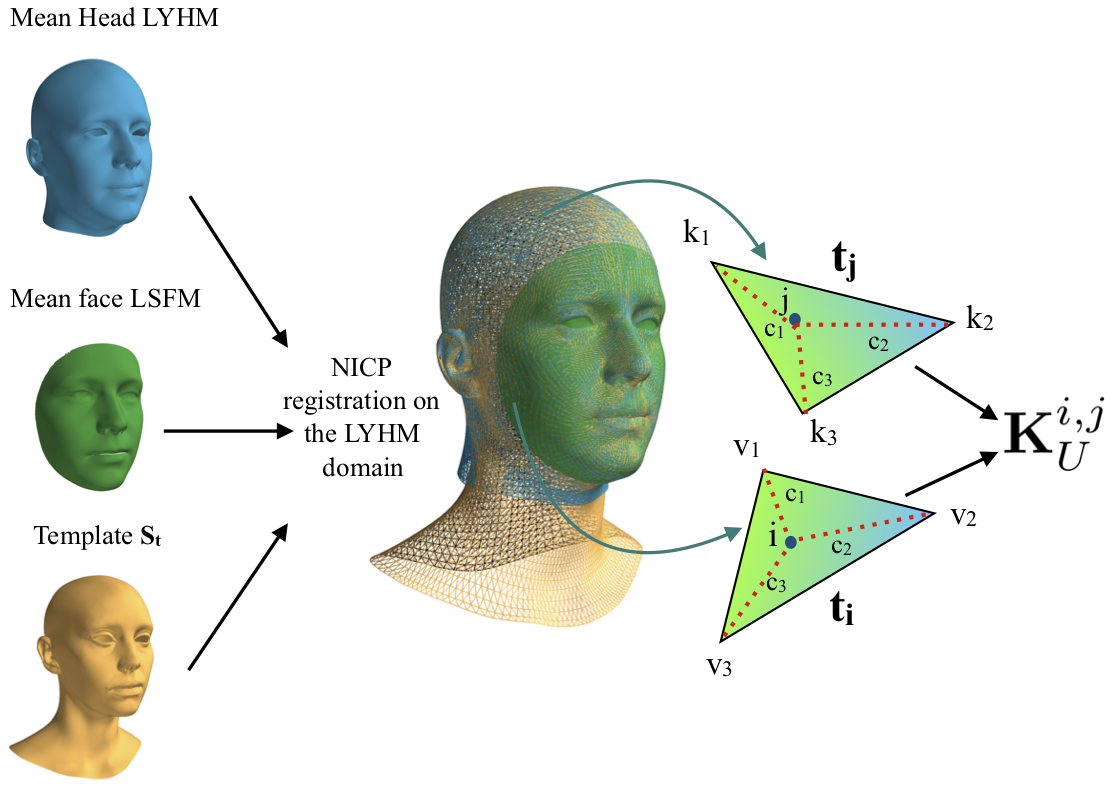}
\caption{A graphical representation of the non-rigid registration of all mean meshes along with our head template $\mathbf{S}_t$ and the calculation of the local covariance matrix $\mathbf{K}_U^{i,j}$ based on the locations of the $i_{th}$ and $j_{th}$ points.}
\label{fig:triangles}
\end{center}
\end{figure}

Each vertex pair $(v,k)$ in between the triangles, has an individual covariance matrix $\mathbf{K}_h^{v,k} \in \mathds{R}^{3\times3}$ with $\mathbf{K}_h^{v,k}\supseteq\mathbf{K}_h$. Therefore, we blend those local vertex-covariance matrices to acquire our final local $\mathbf{K}_U^{i,j}$ as follows:
\begin{equation}
\begin{split}
    \mathbf{K}_U^{i,j} =  \frac{\sum_{v=1}^3\sum_{k=1}^3 w_{v,k}^{i,j} \mathbf{K}_h^{v,k}}{\sum_{v=1}^3\sum_{k=1}^3 w_{v,k}^{i,j}}
\end{split}
\label{equ:blending_cov}
\end{equation}
where $w_{v,k}^{i,j} = \frac{c_v^i + c_k^j}{2}$ is a weighting scheme based on the barycentric coordinates of the $(i,j)$ points. An illustration of the aforementioned methodology can be seen in Figure~\ref{fig:triangles}.

\begin{figure*}[h]
\begin{center}
\includegraphics[width=1\linewidth]{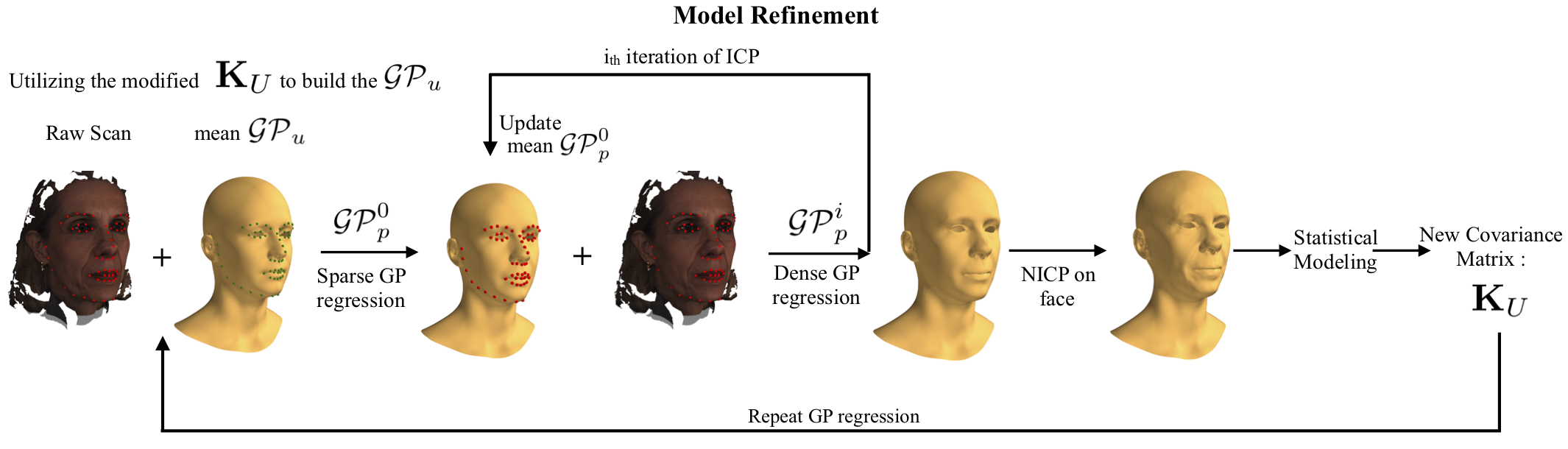}
\caption{The model refinement pipeline. We start with the GP model defined by the universal covariance matrix. For each scan in the \emph{MeIn3D} dataset we obtain full head reconstruction with GP Regression using the sparse landmarks and dense ICP algorithm. We then non-rigidly align the face region of the full head reconstruction to the scan, and build a new sample covariance matrix to update our model.}
\label{fig:intrinsic_exp}
\end{center}
\end{figure*}

In the case where the points lie in the face area, we initially repeat the same procedure by projecting and calculating a blended covariance matrix $\mathbf{K}_f^{i,j}$ given the mean face mesh of LSFM, followed by a blended covariance matrix $\mathbf{K}_h^{i,j}$ calculated given the mean head mesh of LYHM. We formulate the final local covariance matrix as:
\begin{equation}
    \mathbf{K}_U^{i,j} = \rho_{i,j}\mathbf{K}_h^{i,j} + (1-\rho_{i,j})\mathbf{K}_f^{i,j}
    \label{equ:blending_dist}
\end{equation}
where $\rho_{i,j} = \frac{\rho_i + \rho_j}{2}$ is a normalized weight, based on the Euclidean distances $(\rho_i,\rho_j)$ of the $(i,j)$ points from the nose-tip of the registered meshes. We apply this weighting scheme to smoothly blend the head and face models and avoid discontinuities that appear on the borders of the face and head area. 

Lastly, when the points belong to different areas (\ie $i_{th}$ point on face, $j_{th}$ point on head) we simply follow the first method that exploits just the head covariance matrix $\mathbf{K}_h$, since the correlation of the face/head shape only exist in the LYHM. After repeating this methodology for every point pair in $\mathbf{S}_t$ and calculating the entire joined covariance matrix $\mathbf{K}_{U}$, we are able to sample new instances from the Gaussian process morphable model.

\subsection{Model Refinement}
\label{sec:model_ref}
To refine our model, we begin by exploiting the already trained GPMM of the previous section. 
With our head template $\mathbf{S}_{t}$ and the universal covariance matrix $\mathbf{K}_{U}$, we define a kernel function:
\begin{equation}
    k_{U}(\mathbf{x}, \mathbf{y}) = \mathbf{K}_{U}^{CP(\mathbf{S}_{t}, \mathbf{x}), CP(\mathbf{S}_{t}, \mathbf{y})}
\end{equation}
where $\mathbf{x}$ and $\mathbf{y}$ are two given points from the domain where the Gaussian process is defined and the function $CP(\mathbf{S}_{t}, \mathbf{x})$ returns the index of the closest point of $\mathbf{x}$ on the surface $\mathbf{S}_{t}$. We then define our GPMM as:
\begin{equation}
    \mathcal{GP}_{U}(\mu_{U}, k_{U})
\end{equation}
where $\mu_{U}(\mathbf{x}) = [0, 0, 0]^{T}$. For each scan in the MeIn3D dataset, we first try to reconstruct a full head registration with our GPMM using Gaussian Process Regression \cite{luthi2017gaussian, gerig2018morphable}. Given a set of observed deformations $\mathbf{X}$ subject to Gaussian noise $\epsilon \sim \mathcal{N}(0, \sigma^{2})$, Gaussian process regression computes a posterior model $\mathcal{GP}_{p}(\mu_{p}, k_{p}) = posterior(\mathcal{GP}, \mathbf{X})$. The landmark pairs between a reference mesh and the raw scan define a set of sparse mappings, which tells us exactly how the points on the reference mesh will deform. Any sample from this posterior model will then have fixed deformations on our observed points \ie facial landmarks. The mean $\mu_{p}$ and covariance $k_{p}$ are computed as:
\begin{equation}
    \mu_{p}(\mathbf{x}) = \mu(\mathbf{x}) + K_{\mathbf{X}}(\mathbf{x})^{T}(\mathbf{K}_{\mathbf{XX}} + \sigma^{2}\mathbf{I})^{-1}\mathbf{X}
\end{equation}
\begin{equation}
    k_{p}(\mathbf{x}, \mathbf{y}) = k_{u}(\mathbf{x}, \mathbf{y}) - K_{\mathbf{X}}(\mathbf{x})^{T}(\mathbf{K}_{\mathbf{XX}} + \sigma^{2}\mathbf{I})^{-1}K_{\mathbf{X}}(\mathbf{y})
\end{equation}
where 
\begin{equation}
K_{\mathbf{X}}(\mathbf{x}) = (k_{U}(\mathbf{x}, \mathbf{x}_{i})),~ \forall~ \mathbf{x}_{i} \in \mathbf{X}
\end{equation}

\begin{equation}
    \mathbf{K}_{\mathbf{XX}} = (k_{U}(\mathbf{x}_{i}, \mathbf{x}_{j})),~ \forall~\mathbf{x}_{i}, \mathbf{x}_{j} \in \mathbf{X}
\end{equation}

For a scan $\mathbf{S}$ with landmarks $\mathbf{L}_{\mathbf{S}} = \{\mathbf{l}_{1}, ... \mathbf{l}{n}\}$, we first compute a posterior model based on the sparse deformations defined by the landmarks:
\begin{equation}
    \mathcal{GP}_{p}^{0}(\mu_{p}^{0}, k_{p}^{0}) = posterior(\mathcal{GP}_{U}, \mathbf{L}_{\mathbf{S}} - \mathbf{L}_{\mathbf{S}_{t}})
\end{equation}
We then refine the posterior model with Iterative Closest Point algorithm. More specifically, at each iteration $i$ we compute the current regression result as $\mathbf{S}_{reg}^{i} = \{\mathbf{x} + \mu_{p}^{i - 1}(\mathbf{x}) | \mathbf{x}\in \mathbf{S}_t \}$, which is the reference shape wrapped with the mean deformation of the posterior model $\mathcal{GP}_{p}^{i-1}$. We then find the closest points $\mathbf{U}^{i}$ for each point in $\mathbf{S}_{reg}^{i}$ on $\mathbf{S}$, and update our posterior model as:
\begin{equation}
    \mathcal{GP}_{p}^{i+1}(\mu_{p}^{i+1}, k_{p}^{i+1}) = posterior(\mathcal{GP}_{p}^{0}, \mathbf{U}^{i} - \mathbf{S}_{reg}^{i})
\end{equation}
Since the raw scans in the MeIn3D database can be noisy, we exclude a pair of correspondence $(\mathbf{x}, \mathbf{U}(\mathbf{x}) )$ if $\mathbf{U}(\mathbf{x})$ is on the edge of $\mathbf{S}$ or the distance between $\mathbf{x}$ and $\mathbf{U}(\mathbf{x})$ exceed a threshold. After the final iteration we obtain the regression result $\mathbf{S}_{reg} = \{\mathbf{x} + \mu_{p}^{final}(\mathbf{x}) | \mathbf{x}\in \mathbf{S}_t \}$. We then non-rigidly align the face region of $\mathbf{S}_{reg}$ to the face region of the raw scan to obtain our final reconstruction.

In practice, we noticed that the reconstructions often produce unrealistic head shapes. We therefore modify the covariance matrix $\mathbf{K}_{U}$ before the Gaussian process regression. We first compute the principal components by decomposing $\mathbf{K}_{U}$, then reconstruct the covariance matrix using  \eqref{equ:compute_covariance} with fewer statistical components. With the full head reconstructions from the MeIn3D dataset, we then compute a new sample covariance matrix, and repeat the previous GP regression process to refine the reconstructions. Finally we perform PCA on the refined reconstructions to obtain our final refined model.
\section{Ear model combination}
\label{sec:ear_modeling}
Both of the original 3DMMs, LSFM (face) and LYHM (head), exhibit only moderate statistical variation around the ear area. 
In order to overcome this limitation, we augment our combined face and head model by creating a high-resolution model of the ears, constructed from $254$ distinct scans. 

\subsection{High resolution ear model}
To construct a 3DMM from a sufficiently large sample size, we draw on several data sources. As with previous ear models, we make use of the SYMARE database \cite{jin2013creating}, which provides both left and right ears of $10$ individuals. Additionally, we have built a dataset of $121$ distinct high-resolution ears from $64$ individuals (32 males and 32 females) ranging from 20 to 70 years old, by scanning the inner and outer area of both ears with a light-stage apparatus. In order to amplify the statistical variation of our ear model across all ages, we employ an additional $113$ ears of children, acquired via CT scans.

\begin{figure}[h]
    \centering
    \includegraphics[width=1\linewidth]{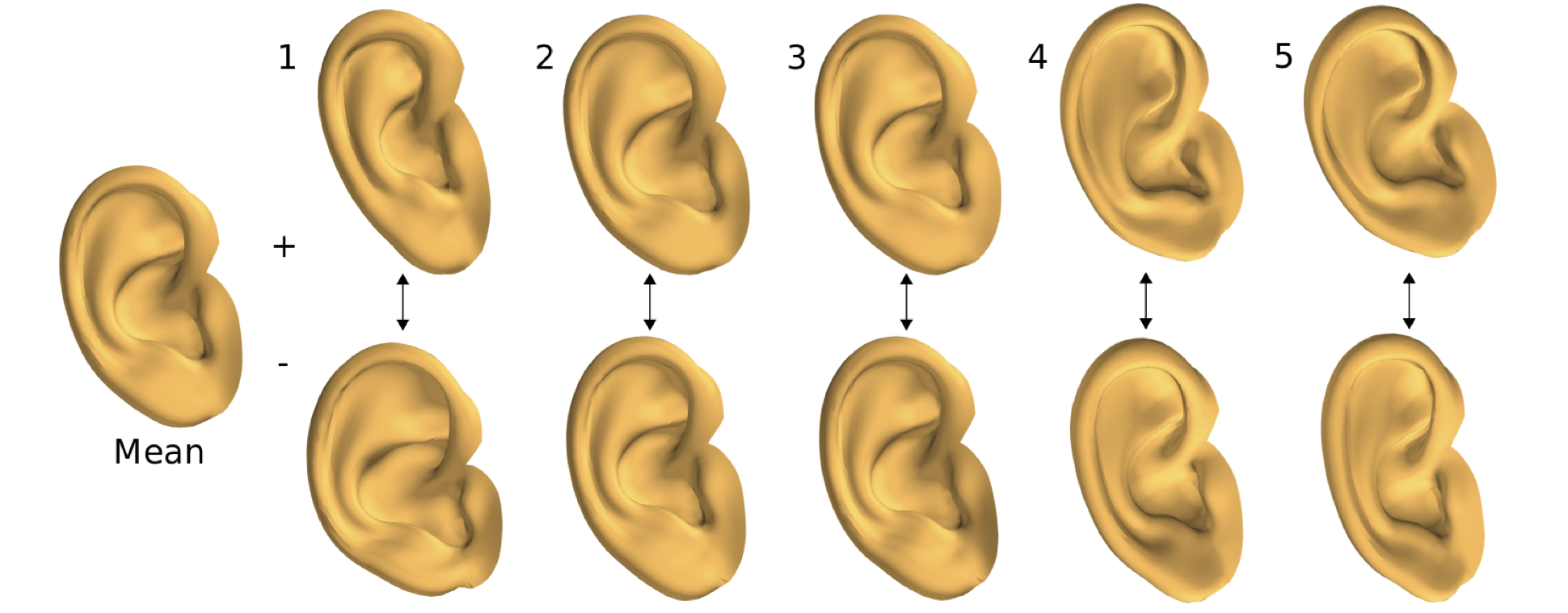}
    \caption{Visualization of the first five principal shape components (with \textpm 3 standard deviations) of our ear model along with the mean ear shape.}
    \label{fig:ear_comp}
\end{figure}

The combined dataset comprises $254$ meshes. All left ears were mirrored to be consistent with the right ears. Each of the meshes was manually annotated with $50$ landmark points to guide the registration process, and then put in correspondence with a template containing $2800$ vertices ($N_e$) using the same NICP-variant non-rigid registration framework employed for the LSFM \cite{booth20163d}. These meshes are then rigidly aligned using Generalised Procrustes Analysis (GPA). Applying PCA to all points in the aligned meshes yields a high resolution 3DMM of the right ear. A 3DMM of the left ear is obtained by reflecting the right ear 3DMM in the sagittal plane, both in terms of its mean shape and its principal components.
The resulting shape components of our right ear 3DMM can be seen in Figure~\ref{fig:ear_comp}.

\subsection{Fusing the ear and the head model}
In order to accurately incorporate new ear shape variations into our combined face and head model, we exploit the same methodology of Gaussian process modeling, as described in Section~\ref{sec:GP_modeling}. We begin by merging, in a non-rigid manner, the mean shape templates ($\mathbf{S}_{le}$, $\mathbf{S}_{re}$) of each ear model (left and right) to the ears of our mean head mesh after the combination of the LSFM (face) and LYHM (head) models. Once all the mean templates are registered we calculate the covariance matrices for each individual model:
\begin{align*}
    \mathbf{K}_{re} &= \mathbf{U}_{re} \mathbf{ \Lambda}_{re}  \mathbf{U}_{re}^T \\
    \mathbf{K}_{le} &= \mathbf{U}_{le} \mathbf{ \Lambda}_{le}  \mathbf{U}_{le}^T
\end{align*}
where $\mathbf{K}_{re} \in \mathds{R}^{3N_e\times 3N_e}$ , $\mathbf{K}_{le} \in \mathds{R}^{3N_e\times 3N_e}$ are the covariance matrices and the $\mathbf{\Lambda}_{re}\in \mathds{R}^{3n_{re}\times 3n_{re}}$ and $\mathbf{\Lambda}_{le}\in \mathds{R}^{n_{le}\times n_{le}}$ are diagonal matrices of eigenvalues for the right and left ear respectively. Our goal is to enhance the ear shape variation of the combined covariance matrix $\mathbf{K}_U$. We begin by merging the right ear model and, as before, we keep the mean head template as a reference. For each projected point pair $i,j$ that belongs in the right ear area, we identify their exact location in the registered $\mathbf{S}_{re}$ mesh in terms of barycentric coordinates with respect to the corresponding triangles. Inbetween each vertex pair, we blend the local covariance matrices, as before, with \eqref{equ:blending_cov}. We then perform the same procedure for the left ear model.

In order to correctly incorporate both ear models into the full head model, we need to introduce a blending distance function that helps avoid discontinuities on the merge borders around the ear base. We adopt \eqref{equ:blending_dist} as the blending mechanism for our covariance matrices and we seek to find a suitable $\rho_{i,j}$ normalized weighing scheme for the points pairs $i,j$ that belong in the ear templates. Naturally, ears form an elongated shape where a Euclidean distance from the base of the ear to the outer parts becomes an unsuitable measure for weighting the point pairs correctly. Instead, we first unwrap the ear mesh into a circle in 2D space, where the center belongs to the ear canal and the furthest points of the circle correspond to the base of the ear. The blending $\rho_{i,j}$ scheme is now measured in the 2D flattened space where distances ${\rho_i,\rho_j}$ are calculated from the center of the unwrapped circle. Essentially, we enrich independently the local variations of each ear without interfering with the already learned variations between them. The resulting fused model is able to describe asymmetrically any possible ear variations.

\section{Eye model combination}
\label{sec:eye_modeling}

Accurate modeling of the characteristics of human eyes, such as gaze direction, pupil size, iris color and eyelid and eye region shape is important for creating realistic 3D face models. Both LSFM and LYHM models include limited variation of the eyelid shape, due to the low resolution of this region in the original scans, while no other characteristics of the eyes are described by these models. 
To overcome these limitations, we model the eyes and peripheral eye regions with separate statistical models that we incorporate in our final head model.

\subsection{Eye models}
\label{subsub:eye_models}
We initially utilize a classical 3DMM optimization framework, under which we combine a statistical model of the eyelid shape and a statistical model of the eye to accurately recover eyelid shape, gaze direction and pupil size from images.

\begin{figure*}[h]
    \centering
    \includegraphics[width=1\linewidth]{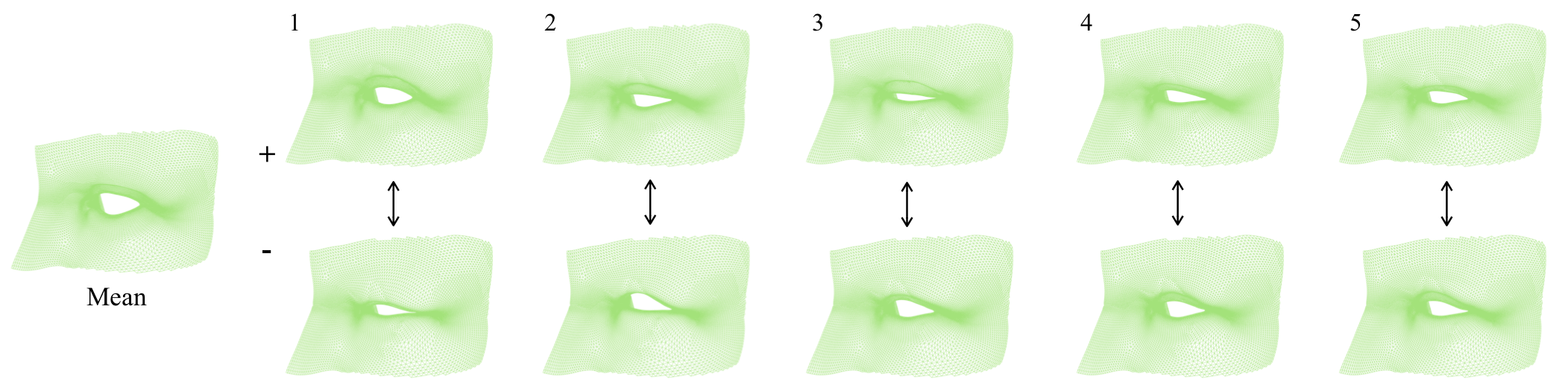}
    \caption{Illustration of the first five components of the eye region shape model $\mathbf{S}_{el}$ that outlines the eyelid shape along with the peripheral shape around the eye. Our model demonstrates large variance and is capable of reconstructing any given eyelid and eye region shape across the human population.(\ie round, almond, monolid, hooded, upturned and downturned eyes)}
    \label{fig:eye_blends}
\end{figure*}

\smallskip
\textbf{Eye region shape model}: To capture the variation of eyelids and the peripheral eye region in the human face, we constructed a PCA model based on $72$ distinct 3D head meshes sculpted around the eye region by a graphics artist. Thus, the eye region shape model can be expressed as:
\begin{equation}
\mathbf{S}_{el} = \mathbf{\bar{s}}_{el} + \mathbf{U}_{el}\mathbf{p}_{el}
\end{equation}
where $\mathbf{\bar{s}}_{el} \in \mathcal{R}^{3N_{U}}$ is the mean shape, $\mathbf{U}_{el} \in \mathcal{R}^{3N_{U} \times M}$ is the eye region shape subspace with dimension $M$ and $\mathbf{p}_{el} \in \mathcal{R}^{M}$ are parameters of the model. The five most significant statistical components of our eye region model are depicted in Figure \ref{fig:eye_blends} cropped in a patch around the left eye area. The variation of our eye region PCA model can accurately represent different types of eye region shapes, such as round eyes, almond eyes, monolid, hooded eyes, upturned and downturned eyes.

\begin{figure*}[h]
    \centering
    \includegraphics[width=0.8\linewidth]{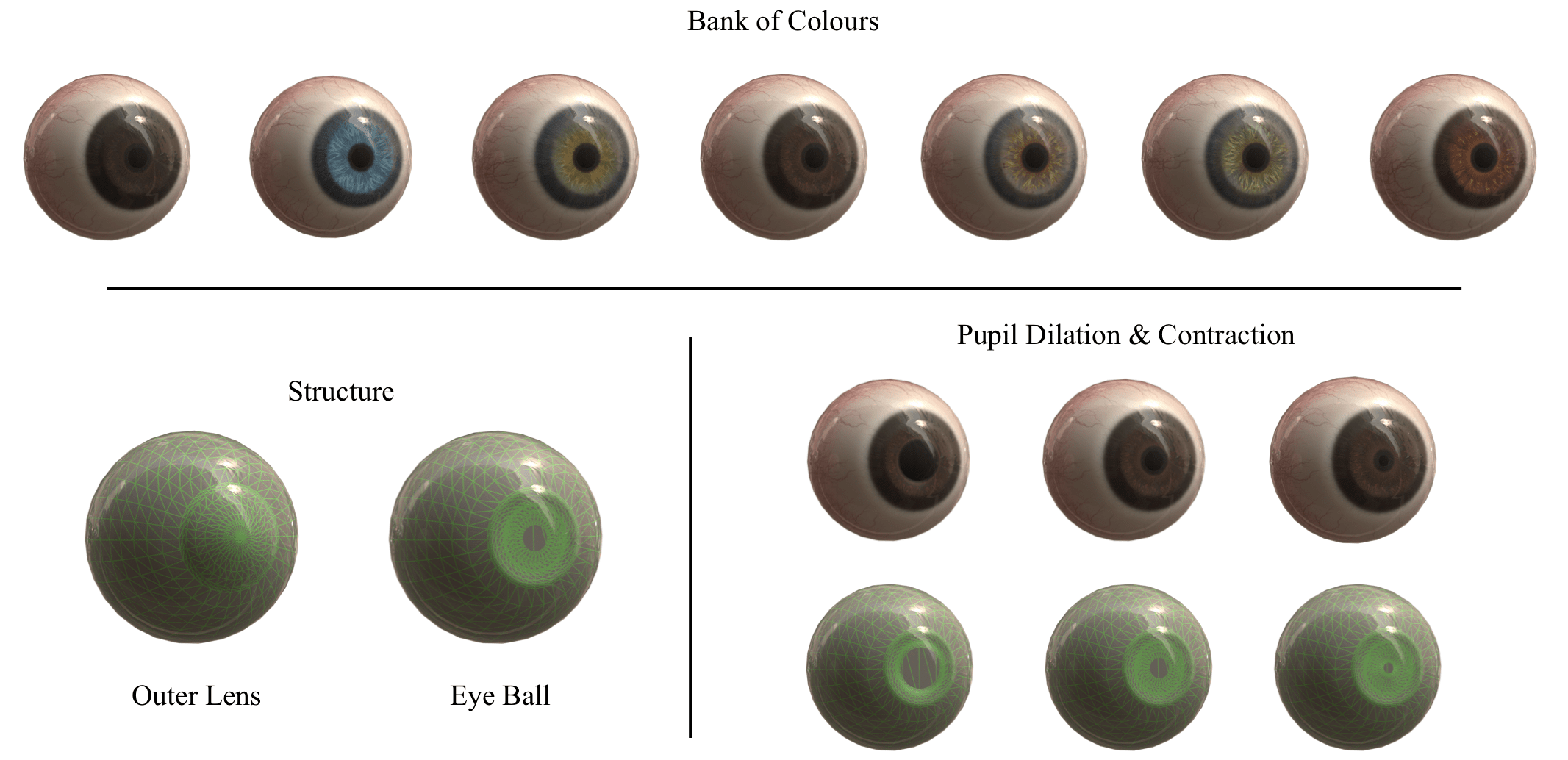}
    \caption{The bank of iris textures in our model along with our eye mesh structure. Our entire eye mesh topology consists of two meshes: The outer one is a transparent mesh that forms the lens and the internal mesh is the eyeball, depicting the iris texture and the pupil shape. On the bottom right corner, we illustrate variations of pupil dilation and contraction with and with out texture.}
    \label{fig:eye_colours}
\end{figure*}

\smallskip
\textbf{Eye shape model}: We model the eye gaze direction and pupil size separately from the eye region. Particularly, we employ two separate meshes to model the eye, the outer lens and the eyeball, as depicted in Figure \ref{fig:eye_colours}. The outer lens covers the eyeball and is static in shape, while the eyeball includes the iris and leaves a hole for the pupil to become visible. To control pupil dilation and constriction, we manually created a blendshape of the pupil size by sculpting an eyeball instance with a different pupil size and subtracting it from the original mesh. The blendshape allows us to render eyeballs with arbitrary pupil size, as in Figure \ref{fig:eye_colours}. For consistency with the eye region model, we express the eye model as a linear combination of the mean eye shape $\mathbf{\bar{s}}_{eye} \in \mathcal{R}^{3N_{eye}}$ and the blendshape $\mathbf{u}_{s, eye} \in \mathcal{R}^{3N_{eye}}$ as $\mathbf{S}_{eye} = \mathbf{\bar{s}}_{eye} + p_{eye}\mathbf{u}_{s, eye}$, where $p_{eye}$ is a parameter of the model and $N_{eye}$ is the number of vertices of the eye model.

\smallskip
\textbf{Eye texture model}: To boost the reconstruction accuracy of our eye model when fitting to
input images, and to recover the color of the eyes along with the shape, we attach an RGB texture model $\mathbf{T}_{eye}$ on our eyeball model $\mathbf{S}_{eye}$, extracted from 2D images. To build $\mathbf{T}_{eye}$ we utilized 100 frontal images of human irises, which we manually annotated with respect to 16 landmarks around the iris and pupil. Then, for each image we projected our eyeball model on the image plane based on 8 iris landmarks and manually adjusted $p_{eye}$ to match the 8 pupil landmarks.
We sampled the images at the projected vertex locations of the iris of our model, to create per-vertex textures. For the locations outside the iris, we used white to represent the sclera of the eye and black to represent the pupil. Finally, we created a PCA model for the per-vertex texture of our 3D eyeball, which can be written as:
\begin{equation}
\mathbf{T}_{eye} = \mathbf{\bar{t}}_{eye} + \mathbf{U}_{t, eye}\mathbf{\lambda}
\label{eq:texture_model_eye}
\end{equation}
where $\mathbf{\bar{t}}_{eye} \in \mathcal{R}^{3N_{eye}}$ is the mean texture component, $\mathbf{U}_{t, eye} \in \mathcal{R}^{3N_{eye} \times M_{eye}}$ is the eye texture subspace with dimension $M_{eye}$ and $\mathbf{\lambda} \in \mathcal{R}^{M_{eye}}$ are parameters of the model.

\begin{figure}[h]
    \centering
    \includegraphics[width=1\linewidth]{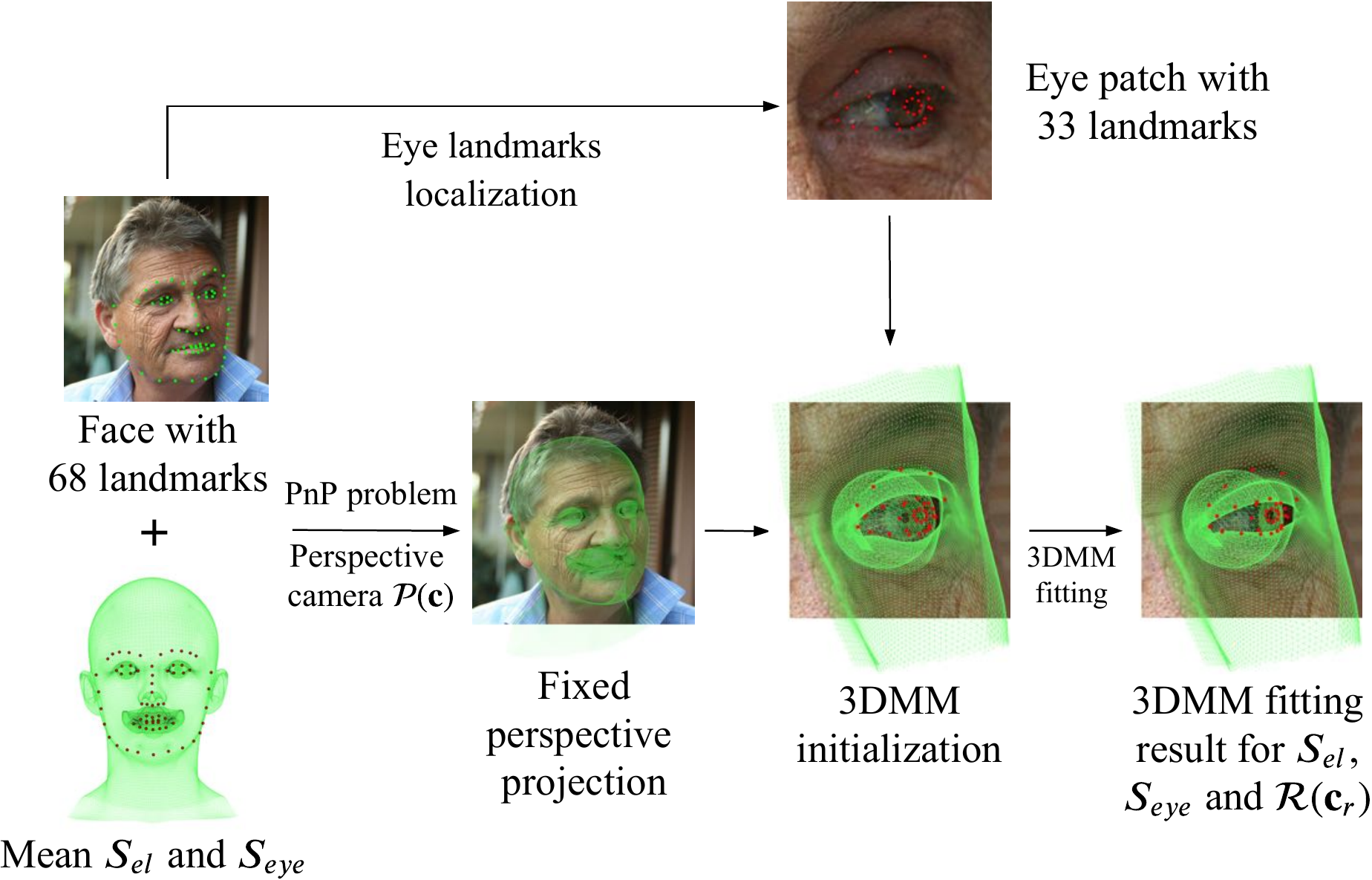}
    \caption{Eye 3DMM fitting pipeline for recovering eye region shape, gaze direction and pupil size from single images.}
    \label{fig:eye_fit}
\end{figure}

\subsection{Optimization-based eye model fitting}
\label{subsub:fitting}
To automatically recover eyelid shape, gaze direction, pupil size and iris color from images, we employ a 3DMM fitting approach in which we optimize our parametric models of shape and texture, based both on 2D landmarks and the texture of the eyes in images. To this end, we automatically extract 33 2D landmarks from images, by utilizing a deep network with hourglass architecture \cite{jiankangfg2018}, which we trained on 3000 images that we manually annotated. The 33 eye landmarks are composed of 17 eyelid landmarks around the eye sclera and on the upper eyelid, denoted as $l_{el}$, and 16 landmarks around the iris and pupil, denoted as $l_{eye}$. 

The fitting pipeline is then split in two steps, as shown in Figure \ref{fig:eye_fit}. First, we recover a perspective camera viewpoint $\mathcal{P}(\mathbf{c})$ for the whole head by solving a Perspective-n-Point (PnP) problem between 68 2D face landmarks of the image, which we extract with \cite{jiankangfg2018}, and 68 3D landmarks of the head model. Then, keeping the head camera fixed, we optimize our statistical eye models based on two landmarks losses and a rendering loss. We model gaze direction as an independent 3D rotation $\mathcal{R}(\mathbf{c}_r)$ relative to the head perspective camera $\mathcal{P}(\mathbf{c})$. In our camera models, vector $\mathbf{c} = [f, t_x, t_y, t_z, q_0, q_1, q_2, q_3]^T$ includes parameters for the focal length, translation and rotation, while vector $\mathbf{c}_r = [q_{r0}, q_{r1}, q_{r2}, q_{r3}]^T$ includes only rotation parameters. In both camera transformations, rotation is modeled with quarternions because of the ease of incorporating them in the optimization in comparison to Euler angles.

We form the following cost function and solve with respect to our models' parameters:
\begin{equation}
\begin{aligned}
    \argmin_{\mathbf{p}_{el}, p_{eye}, \mathbf{c}_r} 
    & \left\lVert \mathcal{W}_{head}(\mathbf{p}_{el}, \mathbf{c}) - l_{el} \right\rVert^2
    \\
    &+ c_l\left\lVert \mathcal{W}_{eye}(p_{eye}, \mathbf{c}, \mathbf{c}_r) - l_{eye} \right\rVert^2
    \\
    &+ c_t\left\lVert \mathbf{I}(\mathcal{W}_{eye}(p_{eye}, \mathbf{c}, \mathbf{c}_r)) - \mathbf{T}_{eye}(\mathbf{\lambda}) \right\rVert^2
    \\
    &+ \ c_{el}\left\lVert \mathbf{p}_{el} \right\rVert^2_{\Sigma_{p_{el}}^{-1}}
    + c_{eye, l}\left\lVert p_{eye} \right\rVert^2_{\Sigma_{p_{eye}}^{-1}}
    + c_{eye, t}\left\lVert \lambda \right\rVert^2_{\Sigma_{\lambda}^{-1}},
\end{aligned}
\label{eq_3DMM_eye_cost}
\end{equation}
where $\mathcal{W}_{head}(\mathbf{p}_{el}, \mathbf{c}) = \mathcal{P}(\mathbf{S}_{el}(\mathbf{p}_{el}), \mathbf{c})$ is the perspective projection of the eye region shape model in the image plane and $\mathcal{W}_{eye}(\mathbf{p}_{eye}, \mathbf{c}, \mathbf{c}_r) = \mathcal{P}(\mathcal{R}(\mathbf{S}_{eye}(p_{eye}), \mathbf{c}_r), \mathbf{c})$ is the independent rotation and perspective projection of the eye shape model in the image plane. 

In \eqref{eq_3DMM_eye_cost}, the first term accounts for the reconstruction of the eye region shape, based on the eyelid landmarks $l_{el}$, while the second term accounts for the reconstruction of the pupil size and gaze direction, based on the iris and pupil landmarks $l_{eye}$. The third term is a texture loss between image $\mathbf{I}$, sampled at the model's projected locations, and our texture model instance $\mathbf{T}_{eye}(\mathbf{\lambda})$. The last three terms are regularization terms, which serve to counter over-fitting and $c_l$, $c_t$, $c_{el}$, $c_{eye, l}$ and $c_{eye, t}$ are weights used to regularize the importance of each term during optimization. Problem \eqref{eq_3DMM_eye_cost} is solved with the simultaneous variation of Gauss-Newton optimization as formulated in \cite{booth20173d}.

\subsection{Extending the traditional approach}
\label{sec:eye_modeling_regression}
The described 3DMM fitting algorithm produces accurate predictions for eye region, pupil size and gaze direction in images, but is relatively slow and requires multiple Gauss-Newton steps to converge. Thus, we attempted to take the traditional 3DMM fitting approach one step further and trained a regression network to estimate the parameters of our 3D models $\{\mathbf{p}_{el}, p_{eye}, \mathbf{c}_r\}$ in a single forward pass. 

To this end, we have utilized the pretrained hourglass network from Section \ref{subsub:fitting} as an encoder and in the last layers we stack a Multi-Layer Perceptron (MLP) architecture resulting in the parameters of our model $\{\mathbf{p}_{el}, p_{eye}, \mathbf{c}_r\}$. The numbers of neurons in each layer of the MLP are (66, 128, 256, 1024, 512, 256, 128, 10), where the last layer represents the concatenation of the five eye region blendshape parameters, the single pupil blendshape parameter, and the four quaternions that describe the rotation of the eyeball. We trained the entire network end-to-end in a supervised fashion with pairs of 2D images of the eye region and the corresponding parameters that we recovered with our 3DMM fitting pipeline. We manually filtered the 3DMM results, discarding any misaligned meshes  before training. To extract the training image pairs we utilized AgeDB \cite{moschoglou2017agedb}, which contains 16488 images of faces of people of various ages. For training and testing our regression model, we split AgeDB into a training set ($90\%$ of the images) and a test set ($10\%$ of the images). Results are presented in Section \ref{subsec:eye_eval}.



\subsection{Color estimation}
High quality eye color reconstruction is difficult to achieve using low resolution images of the eye region with the standard iris PCA model of Section~\ref{subsub:eye_models}. To this end, we treat the problem of eye color reconstruction as a classification problem, given a bank of known iris textures as shown in Figure~\ref{fig:eye_colours}.

In order to make as accurate predictions as possible with respect to the color of the eyes of a subject depicted in an ``in-the-wild'' image, we need thousands of ground-truth eye images annotated with regards to their color. To this end, we utilized AgeDB \cite{moschoglou2017agedb} and we employed the 68 2D landmarks to extract the eye regions for each one of the images. Subsequently, since there are only seven different colors for human eyes, we manually annotated the extracted eye images with one of the following options: \emph{amber}, \emph{blue}, \emph{brown}, \emph{gray}, \emph{green}, \emph{hazel}, or \emph{dark brown}. The cropped eye images were of size $64\times64$.

We used $90\%$ of the AgeDB data \cite{moschoglou2017agedb} for the training process and the rest for testing. We carried out the training utilizing a simple encoder architecture, similar to the one described in \cite{moschoglou20193dfacegan}. The only modification was with respect to the last layer, where the output dimension was changed to seven, to be in accordance with the total number of eye colors. This architecture yielded the best results, with about $92\%$ classification accuracy in the test set. Given that certain eye classes are highly correlated and are even challenging to classify by humans (such as \emph{amber} and \emph{brown} or \emph{gray} and \emph{blue}), the model actually achieves very high accuracy, since the vast majority of misclassifications occur between these groups.
\section{Oral cavity and teeth}
\label{sec:teeth_inner_vavity}
An appropriate and complete representation of the human head should also model the inner mouth cavity and the teeth, in addition to the external characteristics of the human head, as these are often visible in raw images. Correctly capturing the 3D topology of the oral cavity in a single template is a challenging topic due to the lack of 3D data, the challenging non-convex and specular teeth regions, as well as the highly deformable nature of the tongue. 
To progress this aspect, we have incorporated an inner mouth topology, where we model the lining inside the cheeks, the front two thirds of the tongue, the upper and lower gum, and the floor and the roof of the mouth. We treat teeth as separate meshes and we fix their location on top of the gums. The tongue and the teeth are not fitted to any training data and, as such, they do not capture any independent statistical variance. However, the overall scale for all axes, is copied and back-propagated smoothly in a decaying manner from the outer lips to the inner cavity of our head model.

\section{Texture modeling and completion}
\label{sec:texture_modeling_completion}
Instead of modeling the texture space in a low frequency PCA formulation, we employ a GAN architecture \cite{wang2018high} after bringing in correspondence all the textures in a UV domain space. In this way, we are capable of preserving the high frequency skin details and avoiding the blurriness of a PCA model. Our combined data set of textures consists of approximately $10$K facial textures and $1,200$ full head textures from the original LSFM and LYHM respectively. Unfortunately, the textures of the cranium region are unwanted due to the blue latex hair caps that the subjects were instructed to wear during the image capture process.

In order to properly render the head of a given subject, apart from the shape and the facial texture, we need to also successfully visualize the entire head texture. That is, we need to find an elegant way to fill out the missing head texture, given the facial texture. The main problem that arises in this process is the scarcity of ground truth data of full head textures. To address this issue, given the facial textures, we employed a graphics artist to fill out the corresponding missing head textures. In this way, we created an adequate number of face-head texture pairs which we then used to train a pix2pixHD \cite{wang2018high} model to fill out the missing cranium textures.

\begin{figure}[h]
    \centering
    \includegraphics[width=1\linewidth]{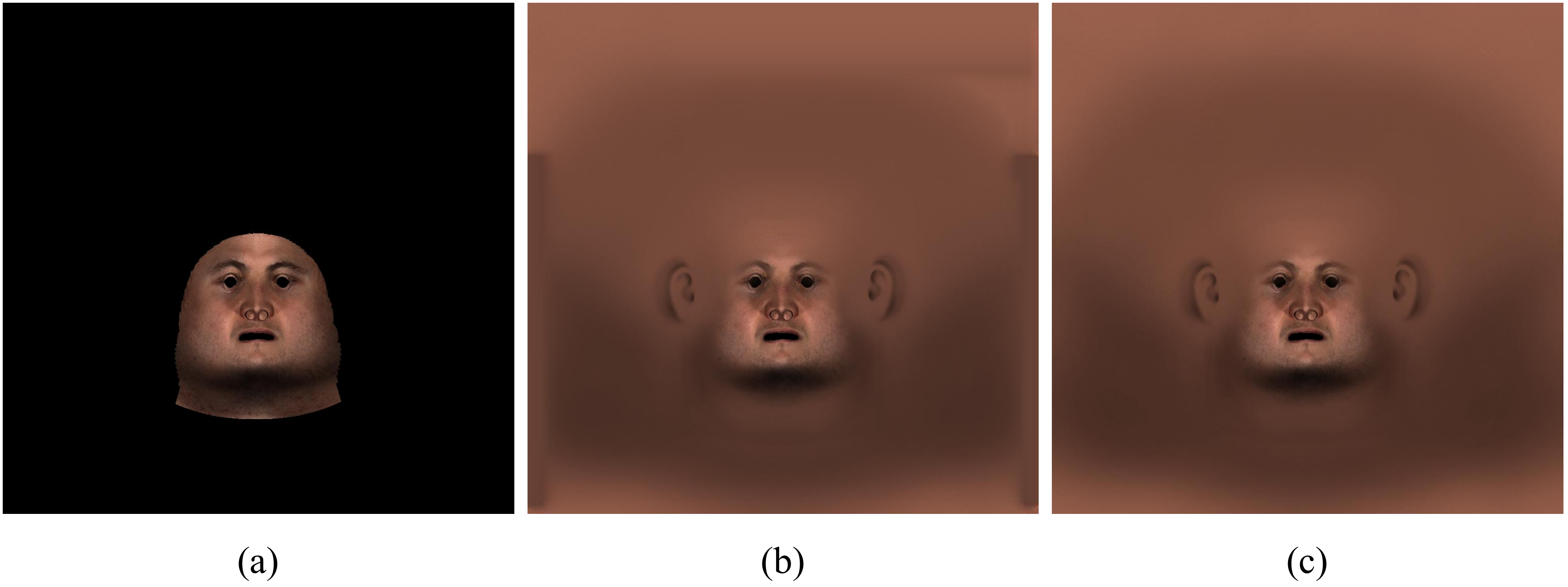}
    \caption{Head texture completion given an unseen facial texture. (a) Input facial texture, (b) recovered completed texture by a pix2pix translation architecture, (c) ground truth completed texture by a graphics artist.}
    \label{fig:texture_completion}
\end{figure}

The pix2pixHD methodology is the current state-of-the-art when it comes to carrying out image translation tasks in high-resolution data. In our case, we learned how to automatically produce complete head textures, given the facial ones. An illustration of a head completion example can be seen in Figure~\ref{fig:texture_completion}. We trained the pix2pixHD model utilizing the learning rates and hyper-parameters mentioned in the original implementation \cite{wang2018high}. However, the global and local blocks in the generator framework were changed to 5 and 10, respectively. Moreover, no instance feature maps were added to the input and, finally, the VGG feature loss was deactivated as this led to a marginally enhanced performance in the completion process.

\section{Experiments}
\label{sec:experiments}
In this section, we analyze in detail the capabilities of our fused head model by examining the intrinsic characteristics in Section \ref{subsec:Intrinsic_eval}. Additionally, in Sections \ref{subsec:head_recon} and \ref{subsec:eye_eval}, we thoroughly describe the full head reconstruction pipeline from 2D images, and we evaluate our approach both qualitatively and quantitatively for all separate attributes.

\subsection{Intrinsic evaluation}
\label{subsec:Intrinsic_eval}
After merging the LSFM face and LYHM head models together, we name our initial head model as the \emph{Combined Face \& Head  Model} (CFHM). When this is augmented into our final model, it is named the \emph{Universal Head Model} (UHM), and this combines four separate models (face, cranium, ears and eyes) into a single representation.

Following common practice, we evaluate our model variations compared to the LYHM by utilizing, \emph{compactness}, \emph{generalization} and \emph{specificity} \cite{davies2008statistical, brunton2014review, bolkart2015groupwise}. For all the subsequent experiments we utilise the original head scans of \cite{dai20173d} from which we have chosen 300 head meshes that were excluded from the training procedure. This test set was randomly chosen within demographic constrains to ensure ethnic, age and gender diversity. We name our model variations as: CFHM-reg built by the regression method, CFHM-GP built by the Gaussian processes kernels framework and finally, CFHM-ref built after refinement with Gaussian process regression. Also, we present bespoke modes in terms of age and ethnicity, constructed by the Gaussian processes kernels method coupled with refinement.

\begin{figure*}[h]
    \centering
    \includegraphics[width=1\linewidth]{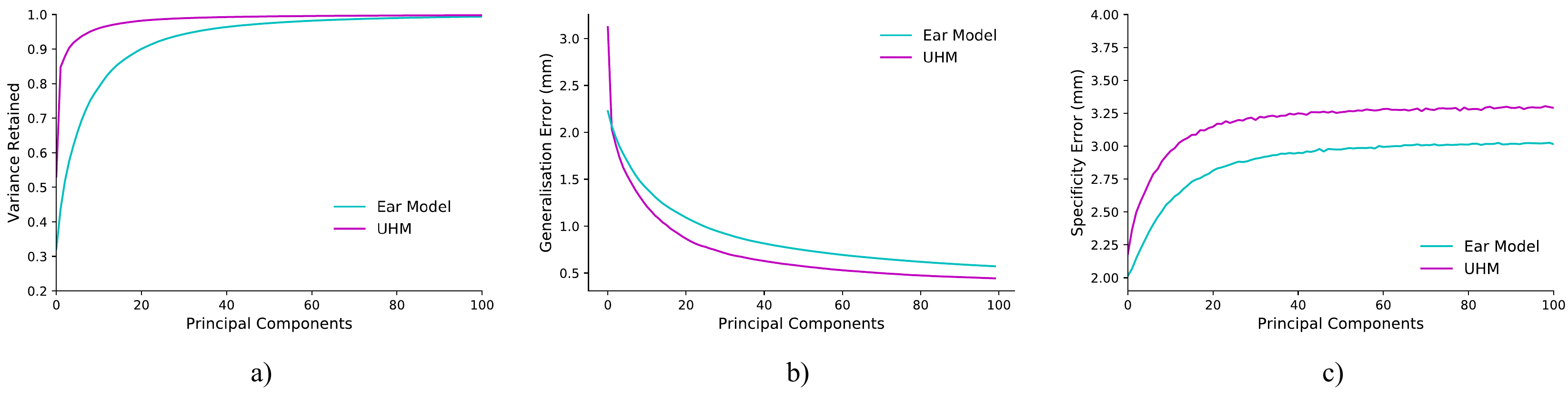}
    \caption{Intrinsic characteristics of the Universal head model (UHM) along with the ear model. a) compactness, b) generalization, c) specificity.}
    \label{fig:intrinsic_uhm_ear}
\end{figure*}

 The top graphs in Figure~\ref{fig:intrinsic_exp} present the compactness measures of the CFHM models compared to LYHM. Compactness calculates the percentage of variance of the training data that is explained by the model, when certain number of principal components are retained. The models CFHM-reg, CFHM-GP express higher compactness compared to the model after the refinement. The compactness ability of the all proposed methods is far greater than the LYHM, as can be seen by the graph. Both global and bespoke CFHM models can be considered sufficiently compact. In Figure~\ref{fig:intrinsic_uhm_ear} (a) the UHM model demonstrates similar compactness to CFHM-reg, CFHM-GP models while extending the variation in the ear area. Compared to the original ear model, the universal model is able to describe the same ear variability with fewer components.

 The center row of Figure~\ref{fig:intrinsic_exp} illustrates the generalization error, which demonstrates the ability of the models to represent novel head shapes that are unseen during training. To compute the generalization error for a given number of principal components retained, we compute the per-vertex Euclidian distance between every sample of the test set and its corresponding model projection and then take the average value over all vertices and test samples. All of the proposed models exhibit far greater generalization capability compared to LYHM. The refined model CFHM-ref tends to generalize better than the other approaches, especially in the range of $20$ to $60$ components. LYHM holds the worst error variance throughout the components whereas the refined model retains the smallest error variance compared to the other models especially in the interval of $40$ to $65$ components. Equivalently, as can be seen in Figure~\ref{fig:intrinsic_uhm_ear} (b), the UHM performs marginally better, but more importantly exhibits regular descent errors compared to all other methods, which ensures stability across all components.
 
 Additionally, we plot the generalization error of the bespoke models against the CFHM-ref in  Figure \ref{fig:intrinsic_exp} (b) center. In order to derive a correct generalization measure for the bespoke CFHM-ref, for every mesh we use its demographic information, we project it on the subspace of the corresponding bespoke model and then we compute an overall average error. We observe that the CFHM-ref mostly outperforms the bespoke generalization models, which might be attributed to the fact that many of the specific models are trained from smaller cohorts, and so run out of interesting statistical variance. This also explains the smaller error variance of the global model throughout the components when compared to the errors of the bespoke models.
 
Finally, the graphs of Figure~\ref{fig:intrinsic_exp} (bottom) show the specificity measures of the introduced models that evaluate the validity of the synthetic faces generated by a model. We randomly synthesize 5,000 faces from each model for a fixed number of components and measure how close they are to the real faces based on a standard per-vertex Euclidean distance metric. We observe that the model that has the best error results is the proposed refined model CFHM-ref. The LYHM model demonstrates better specificity error than the  CFHM-reg, CFHM-GP models only in the first 20 components. Both of the proposed combined models exhibit steady error measures ($\approx 3.8$) after keeping components greater than 20. This is due to the higher compactness that both combined models demonstrate, which enables them to maintain certain specificity error after the 20 components. For all bespoke models, we observe that the specificity errors attain particularly low values, in the range of $1$to $4$ mm. This is evidence that the synthetic head generated by both global and bespoke CFHM models are realistic enough. Additionally, as shown in \cite{booth20163d} the significant error differences between different age groups and ethnicities (\ie black model), are caused by a lack of sufficient representative training data. This means that the training data of those models are insufficient to synthesize new identities, so the nearest neighbor error tends to be greater, as compared to other models with more training examples. Similarly, in Figure~\ref{fig:intrinsic_uhm_ear} (c) the UHM model demonstrates identical specificity measures with CFHM-ref, since the ear fusion does not interfere with the overall ability of the model to synthesize realistic head shapes.

Our results show that our combination techniques yield models that are capable of exhibiting improved intrinsic characteristics compared to the original LYHM head model.

\begin{figure}[h]
    \centering
    \includegraphics[width=1\linewidth]{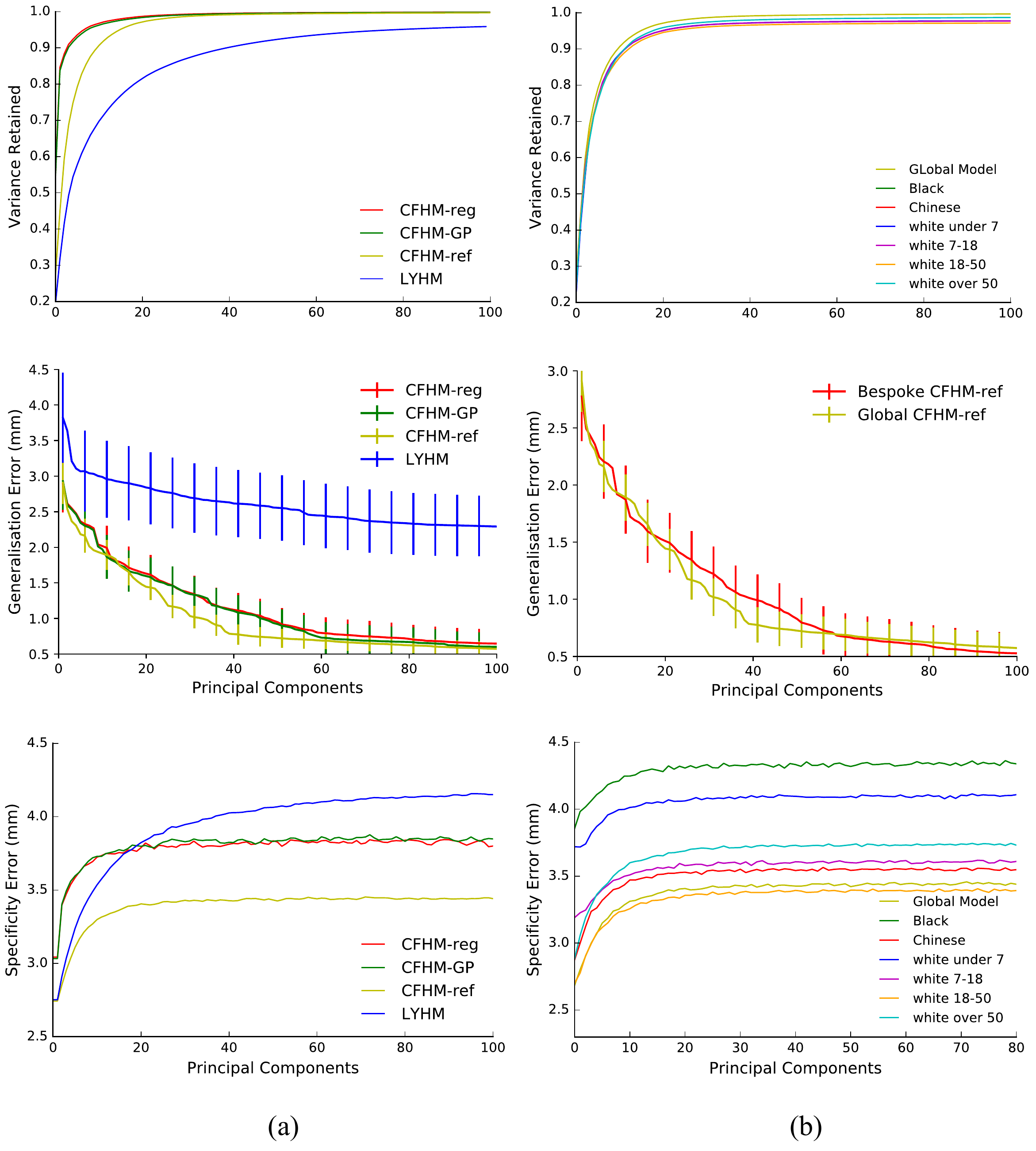}
    \caption{Characteristics of the CFHM models compared to LYHM. Top: compactness; Center: generalization; Bottom: specificity. Left column (a): different methods, Right column (b): demographic-specific 3DMMs based on the CFHM-ref model.}
    \label{fig:intrinsic_exp}
\end{figure}

\begin{figure*}
    \centering
    \includegraphics[width=0.75\linewidth]{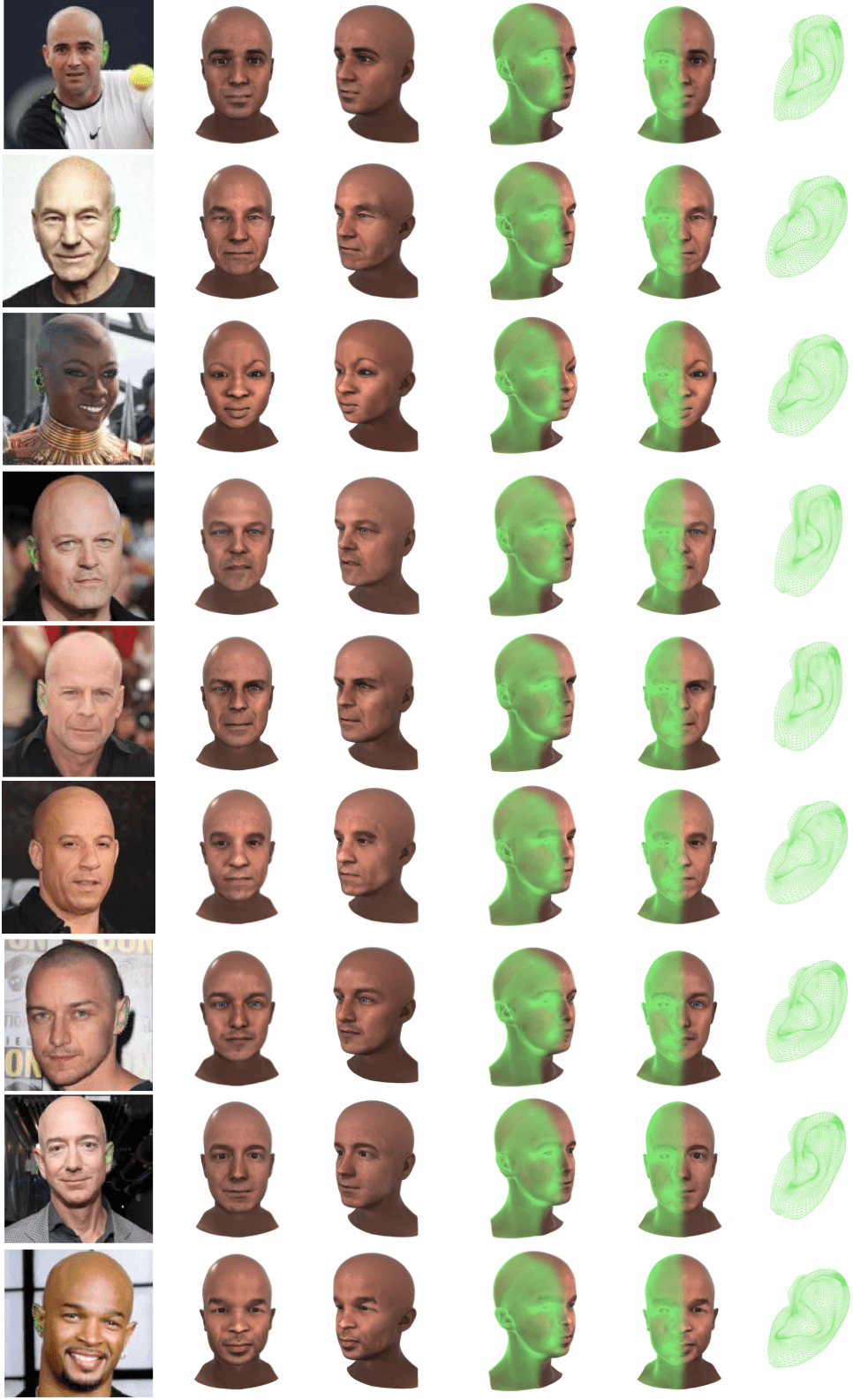}
    \caption{Qualitative results of our in-the-wild 3D head reconstruction. In the first column we show the 2D ear landmarks on top of the subject images. In columns 2-5 we depict our high detailed head reconstruction. In the last column we illustrate different ear shape reconstructions for each subject. Our model is able to generate realistic representations for all general traits (\ie face, head, eyes, ears).}
    \label{fig:heads_qual}
\end{figure*}
\subsection{Head reconstruction from single images}
\label{subsec:head_recon}
By leveraging the UHM model, we outline a methodology that enables us to reconstruct the entire head shape including ears and eye gaze/color from unconstrained single images. We strictly utilize only one view/pose for head reconstruction in contrast to \cite{liang2016head} where multiple images were utilized. We achieve this by regressing from a latent space that represents the 3D face and ear shape to the latent space of the full head models constructed by the proposed methodologies.

We begin by building a PCA model of the inner face along with $50$ landmarks on each ear as described in \cite{zhou2017deformable}. We utilize the $10,000$ head meshes produced by our proposed methods. After building the face-ear PCA model, we project each one of the face-ear examples to get the associated shape parameters $\mathbf{p}_{e/f}$. Similarly, we project the full head mesh of the same identity to the full head PCA model in order to the acquire the latent shape parameters of the entire head $\mathbf{p}_h$. As in Section \ref{reg_combination}, we construct a regression matrix in the same manner, which works as a mapping from the latent space of the ear/face shape to the full head representation.

\begin{figure}[h]
    \centering
    \includegraphics[width=0.9\linewidth]{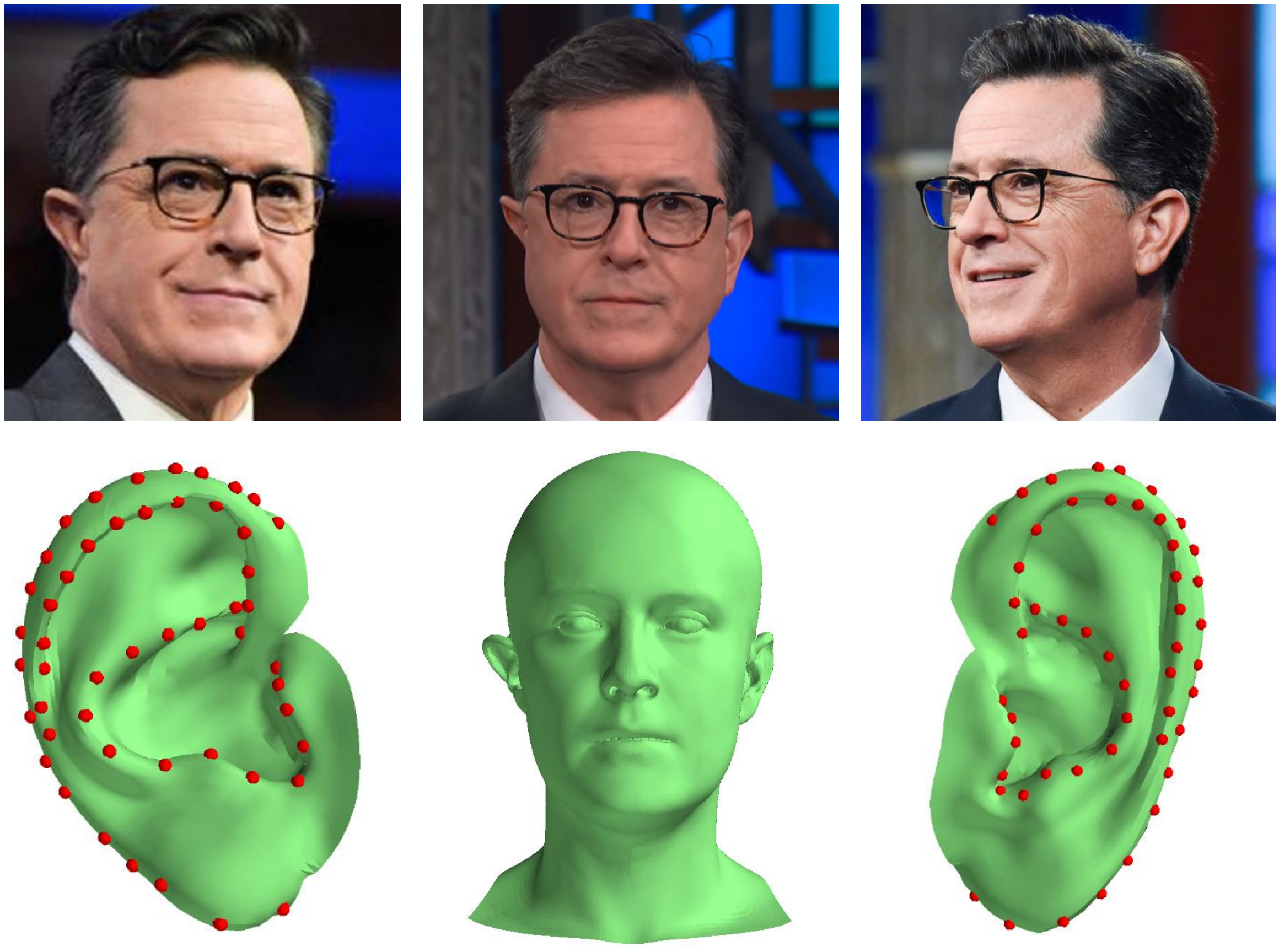}
    \caption{Asymmetric ear reconstruction from different view points for the same identity after retrieving different sets of 3D ear landmarks from the top left and top right images.}
    \label{fig:dif_ear}
\end{figure}

In order to reconstruct the full head shape and texture from 2D images, we begin by fitting the facial part of our head model. Due to the nature of our high frequency head texture model we employ the recently proposed approach in \cite{gecer2019ganfit} where high quality texture reconstructions are possible by leveraging a GAN texture model in a gradient descent optimization setting. Afterwards, we implement an ear detector and an Active Appearance Model (AAM) as proposed in \cite{zhou2017deformable} to localize the ear landmarks in the 2D image domain. Since we have fitted a facial 3DMM in the image space, we already know the camera parameters, \ie, focal length, rotation, translation. To this effect, we can easily retrieve the ear landmarks in the 3D space by solving an inverse perspective-n-point problem \cite{lepetit2009epnp} given the camera parameters and the depth values of the fitted mesh. We mirror the 3D landmarks with respect to the z-axis to obtain the missing landmarks of the occluded ear. After acquiring the facial part and the ear landmarks we are able to attain the full head representation with the help of the regression matrix. Since each proposed method estimates a slightly different head shape for the $10,000$ face scans, we repeat the aforementioned procedure by building bespoke regression matrices for each head model. In order to fill out the entire head texture we employ the texture completion methodology as described in Section~\ref{sec:texture_modeling_completion} where from a facial texture we are able to fill out the entire head surface. Finally, after acquiring the full head shape we refine eye region shape and estimate the eye gaze/color and pupil dilation/contraction, by employing the regression network where the parameters of the eye model are estimated from a cropped image around the eye region. Qualitative results of our approach can be seen in Figure~\ref{fig:heads_qual}. Because of the nature of our complete head model we are able to recover large ear variations among the reconstructed subjects as well as different eye region and head shapes in combination with high quality texture. Asymmetrical ear reconstructions for the same identity are possible when both ears are visible in a multi-view setting as it can be seen in Figure~\ref{fig:dif_ear}.

The overall processing time of the entire head reconstruction pipeline is approximately $32$ seconds on moderate GPUs (\ie NVIDIA RTX 2080 Ti) with CPU Intel Core $i7$ 3.8 GHz. The first part of facial reconstruction along with the ear landmark localization takes around $12$ seconds while the rest of the pipeline that completes the head shape and the texture is processed approximately in $16$ seconds. Finally the eye gaze and colour estimation and the eye region shape along with the pupil dilation/contraction is computed around in $5$ seconds.

\begin{figure*}[h]
    \centering
    \includegraphics[width=1\linewidth]{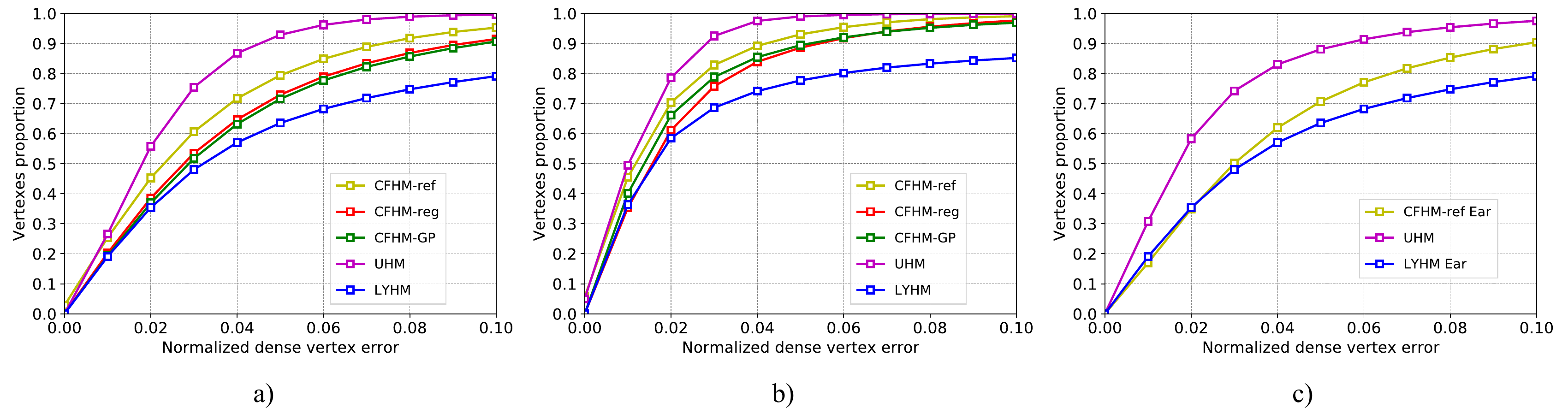}
    \caption{Accuracy results for head and ear shape estimation, as cumulative error distributions of the normalized dense vertex errors. a) accuracy results based on the fitted facial meshes to rendered images, b) accuracy results based on the actual ground truth 3D facial meshes, c) accuracy results only for the ear topology based on the actual ground truth 3D ear meshes. Tables \ref{tab:dense_fit_error_1}, \ref{tab:dense_fit_error_2} and \ref{tab:dense_fit_error_3} report additional measures.}
    \label{fig:fit_exp}
\end{figure*}

We evaluate quantitatively our methodology by rendering $50$ distinct head scans from our test set in frontal and side poses varying from $20$ to $-20$ degrees around the $y$-axis in order for the ears to be visible in the image space. We apply our previous procedure, where we fit a facial 3DMM and we detect the ear landmarks in the image plane. Then for each method we exploit the bespoke regression matrix to predict the entire head shape. We measure the per-vertex error between the recovered head shape and the actual ground-truth head scan by projecting each point of the fitted mesh to the ground-truth and measuring the Euclidean distance. Figure~\ref{fig:fit_exp} shows the cumulative error distribution for this experiment, for the four models under test. Table \ref{tab:dense_fit_error_1} and \ref{tab:dense_fit_error_2} report the corresponding Area Under Curve (AUC) and failure rates for the fitted and the actual ground truth 3D facial meshes respectively. The failure rate, represents the frequency with which a method fails to represent a 3D head shape for a specific amount of vertices within a threshold (the bins of our graph) divided by the total number of bins. Essentially, it is the probability of failure given a threshold.

\begin{figure*}[h]
    \centering
    \includegraphics[width=0.9\linewidth]{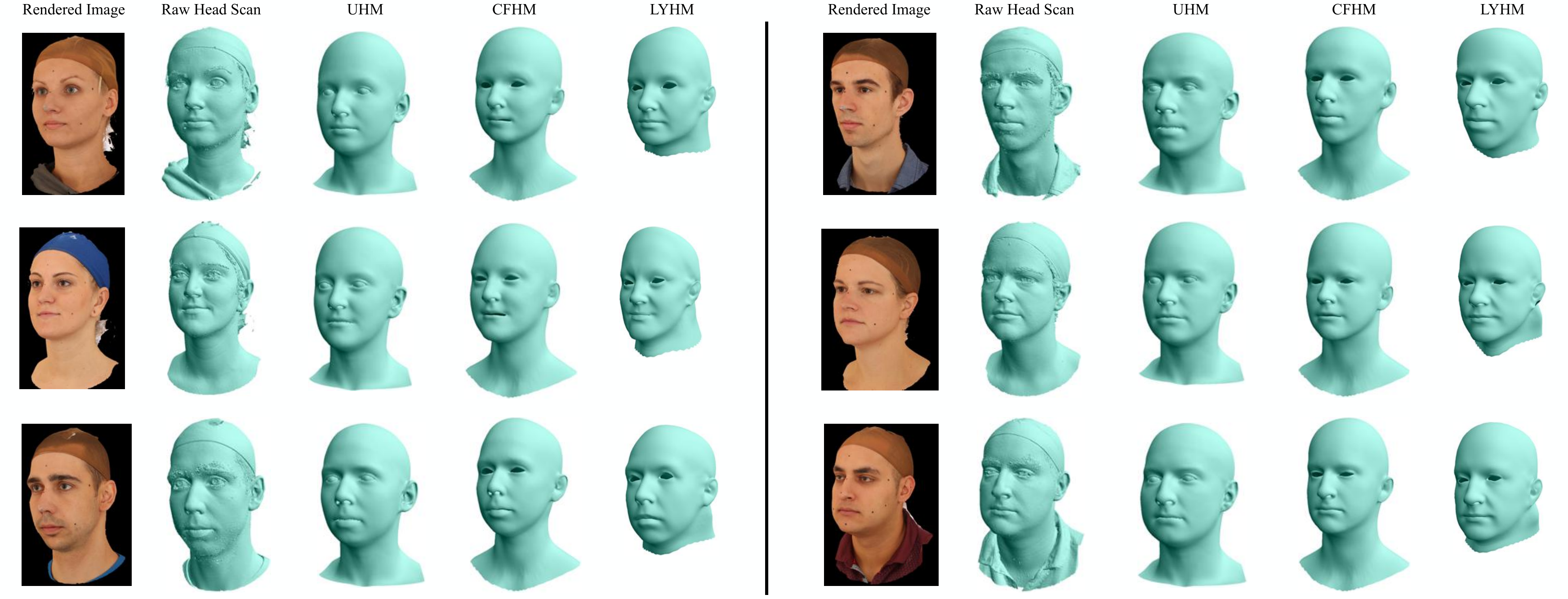}
    \caption{Comparisons of 3D head reconstructions from single images between all head models. From left to right: Rendered textured scan, Original head scan, UHM, CFHM and LYHM reconstructions. Our final universal model is able to reconstruct in detail the entire head shape including the eyes and the ears.}
    \label{fig:qual_ablation}
\end{figure*}

In both situations, the LYHM struggles to recover the head shapes. CFHM-reg and CFHM-GP perform equally, whereas the model after refinement attains better results. The model that exhibits the best reconstruction in both settings is the UHM as shown in the diagrams (a) and (b) of Figure~\ref{fig:fit_exp}. That is attributed to the high quality ear variation of the UHM model after fusion, which the head reconstruction pipeline relies on. By merging the ear model, the degrees of freedom by which the ear topology can drive the entire head shape have significantly increased. Figure~\ref{fig:fit_exp} (c) show the ear shape estimation results of UHM model against the LYHM and the CHFM-ref, from 2D image landmarks compared to the actual ground truth 3D ear meshes. The UHM significantly outperforms both models by a large margin. Additional measures are described in table \ref{tab:dense_fit_error_3}. Some qualitative results can be seen in Figure~\ref{fig:qual_ablation} where we can easily deduce that the UHM outperforms all models in terms of ear estimation and general head shape reconstruction. Both CFHM and LYHM struggle with the prediction of the ear shape because they share the same shape variations. Additionally, LYHM tends to predict arbitrary neck circumferences as well as head shapes, which are excluded from the CHFM and the UHM after model refinement (Section~\ref{sec:model_ref})

\begin{table}[!t]
\centering
\begin{tabular}{|l|ccc|}
\hline
\emph{Method} & \emph{AUC} & Std & \emph{Failure Rate (\%)} \\
\hline\hline
\textbf{UHM} & \textbf{0.875} & \textbf{1.74} & \textbf{1.51} \\
CFHM-ref & 0.751 & 3.42 & 3.64 \\
CFHM-reg & 0.693 & 4.71 & 6.88 \\
CFHM-GP & 0.681 & 4.36 & 7.55 \\
LYHM \cite{dai20173d} & 0.605 & 20.95 & 19.21 \\
\hline
\end{tabular}
\vspace{0.2cm}
\caption{Head shape estimation accuracy results for the fitted facial meshes of our test set. Metrics are Area Under the Curve (AUC), standard deviation (Std) and Failure Rate of the Cumulative Error Distributions of Figure~\ref{fig:fit_exp}.}
\label{tab:dense_fit_error_1}
\end{table}

\begin{table}[!t]
\centering
\begin{tabular}{|l|ccc|}
\hline
\emph{Method} & \emph{AUC} & Std & \emph{Failure Rate (\%)} \\
\hline\hline
\textbf{UHM} & \textbf{0.912} & \textbf{1.12} & \textbf{0.44} \\
CFHM-ref & 0.880 & 2.04 & 0.62 \\
CFHM-GP & 0.844 & 2.81 & 2.46 \\
CFHM-reg & 0.831 & 2.74 & 1.69 \\
LYHM \cite{dai20173d} & 0.739 & 18.14 & 14.10 \\
\hline
\end{tabular}
\vspace{0.2cm}
\caption{Head shape estimation accuracy results for the actual ground truth 3D facial meshes of our test set. Metrics are AUC, standard deviation (Std) and Failure Rate of the Cumulative Error Distributions of Figure~\ref{fig:fit_exp}.}
\label{tab:dense_fit_error_2}
\end{table}

\begin{table}[!t]
\centering
\begin{tabular}{|l|ccc|}
\hline
\emph{Method} & \emph{AUC} & Std & \emph{Failure Rate (\%)} \\
\hline\hline
\textbf{UHM} & \textbf{0.802} & \textbf{2.4} & \textbf{0.32} \\
CFHM-ref Ear & 0.697 & 6.88 & 2.75 \\
LYHM Ear \cite{dai20173d} & 0.621 & 17.52 & 12.6 \\
\hline
\end{tabular}
\vspace{0.2cm}
\caption{Ear shape estimation accuracy results for the ground truth 3D meshes of our test set around the ear area. Metrics are AUC, standard deviation (Std) and Failure Rate of the Cumulative Error Distributions of Figure~\ref{fig:fit_exp}.}
\label{tab:dense_fit_error_3}
\end{table}

\subsubsection{Special Cases}
\begin{figure}[h]
    \centering
    \includegraphics[width=1\linewidth]{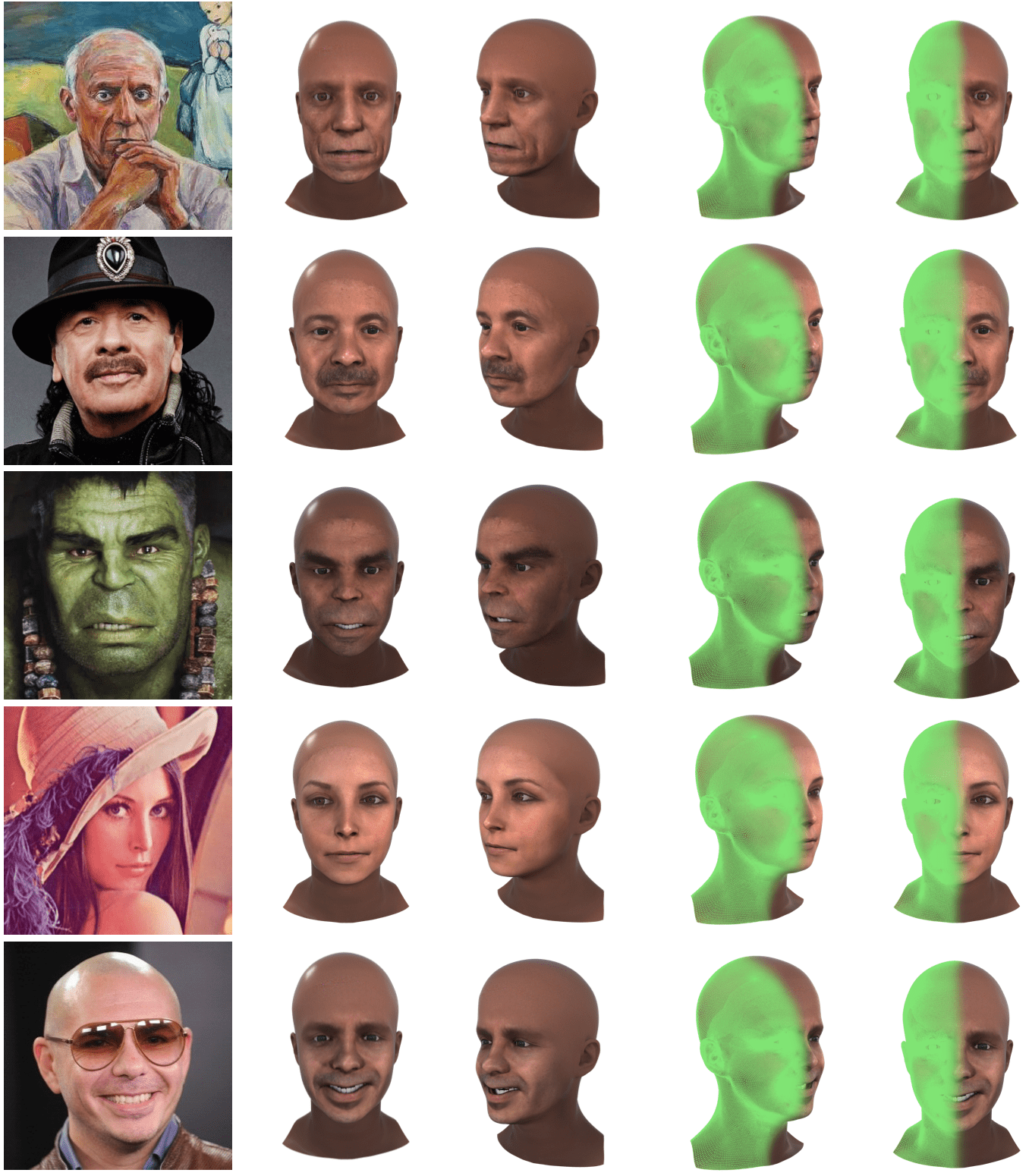}
    \caption{Qualitative results of our 3D head reconstruction pipeline from challenging images with occlusions (hats, sunglasses, hair), images with animated human-like characters (hulk) and images with paintings of people (oil painting of Picasso).}
    \label{fig:special_heads_qual}
\end{figure}

Naturally, in-the-wild faces of people often come with all sorts of occlusions including long hair, hats, sunglasses or even other body parts such as hands covering parts of the face/head. Similar to \cite{gecer2019ganfit}, we rely on a strong optimization setting in order to overcome these limitations. Due to the high frequency nature of our texture model, we are able to exclude any occluding artifacts that might appear and generate realistic head shapes. Also thanks to the face recognition component (face identity features) of \cite{gecer2019ganfit} in the gradient descent optimization framework, we are capable of reconstructing realistic human-like head shapes from oil-paintings and animated characters. As can be seen in Figure~\ref{fig:special_heads_qual}, we are able to reconstruct pleasing head shapes and textures from images with various occlusions (hair, sunglasses, hats, hands) from painting-like images of people and from images of animated characters. In cases where both ears were not visible in the images, we utilized the mean ear landmarks of the UHM model in order to acquire the entire head shape.

\subsection{Eye model evaluation}
\label{subsec:eye_eval}

\begin{figure*}[h]
    \centering
    \includegraphics[width=0.9\linewidth]{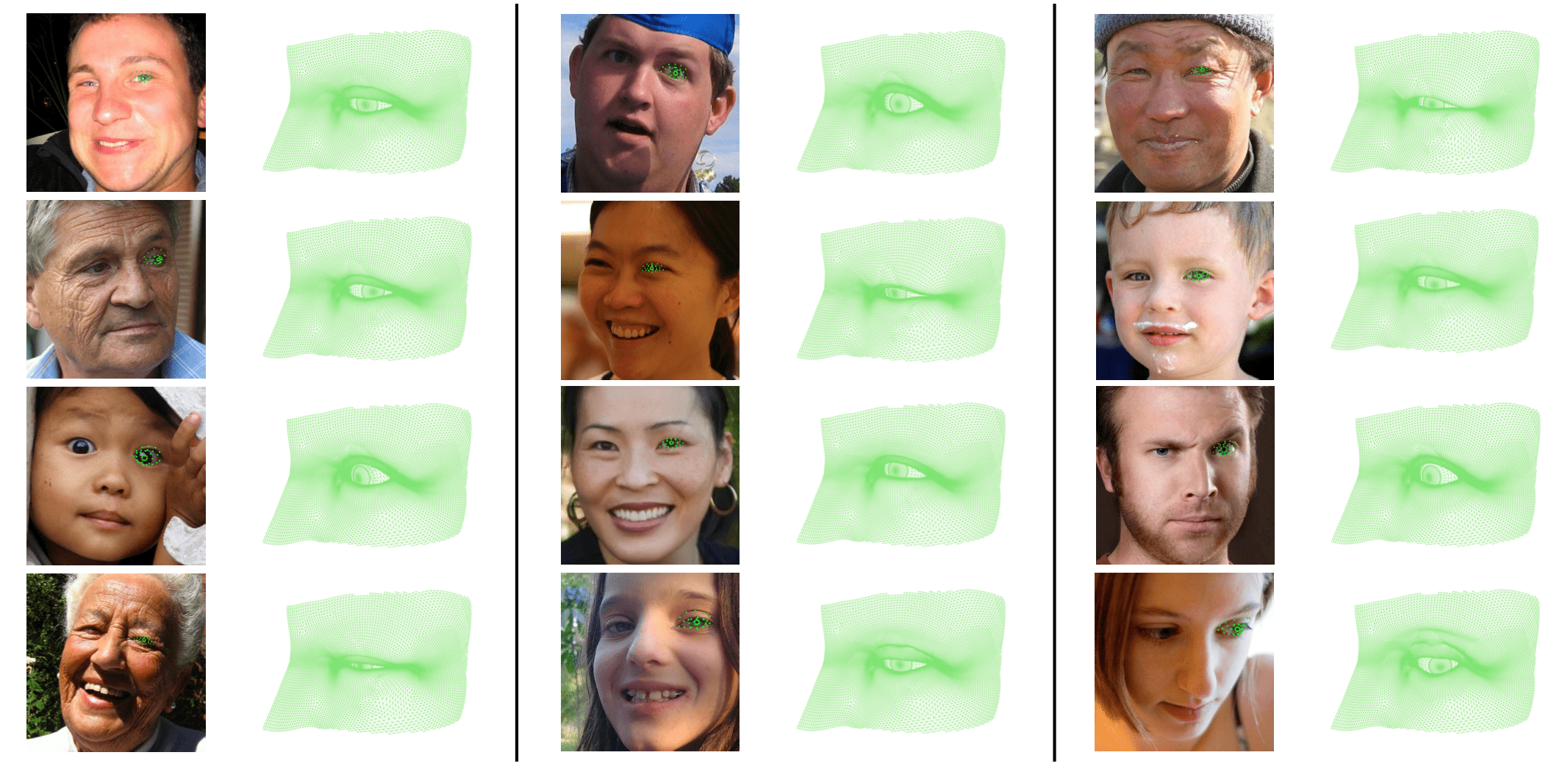}
    \caption{Qualitative results of our eye 3DMM which combines eye lid/outer eye shape reconstruction and gaze/pupil estimation. Our model is capable of reconstructing the correct eye shape and gaze across all possible scenarios (\ie wide one eyes, monolid eyes, upturned eyes, closed eyes).}
    \label{fig:eye_qual}
\end{figure*}

We evaluate the eye modeling pipeline of our UHM both qualitatively in terms of resemblance between reconstructions and input images and quantitatively in the task of gaze estimation from single images. Figure~\ref{fig:eye_qual} includes qualitative results on reconstruction of the eye region from single images by our regression network described in Sec. \ref{sec:eye_modeling_regression}. Reconstructions produced by our pipeline accurately simulate the eyelid shape and gaze direction of the corresponding images, while the pupil size also reasonably adapts to the pupil, wherever it is visible. 

We evaluate our regression network on the testing set of AgeDB \cite{moschoglou2017agedb}, separately for the three regression tasks: eye region shape, gaze direction and pupil size parameters prediction. In more detail, we measure our regression model's accuracy in terms of the Euclidean distance between the the predicted parameters and the ground truth ones which are recovered by our 3DMM pipeline. Furthermore, for comparison reasons we normalise the errors by the $l_2$ norm of the ground truth parameters. To get the total error for each regression task, we average all the sample errors. Finally, our regression network achieved $94\%$ mean parameter accuracy in recovering the eye region shape, $98\%$ mean parameter accuracy in recovering the gaze direction and $86\%$ mean parameter accuracy in recovering the pupil size. 

To further evaluate the eye modeling module of the UHM, we perform a gaze estimation experiment and compare our results with eye3DMM \cite{Wood2016A3M}, in which gaze direction is also estimated by fitting a 3DMM of the eye region. Our model, builds on a similar pipeline and extends it by training an end-to-end network which regresses the 3DMM parameters of our eye models. Table \ref{tab:gaze_stimation} includes gaze estimation results in terms of mean angular errors, on the Eyediap database \cite{FunesMora_ETRA_2014}. For fair comparison with other methods, we didn't include the extreme gaze directions of Eyediap in our experiments. Our model outperforms eye3DMM \cite{Wood2016A3M} by $0.59^o$.

\begin{table*}[!t]
\centering
\begin{tabular}{|l|c|c|c|c|c|c|c|c|}
\hline
 & \emph{\textbf{ours}} & Eye3DMM & CNN & RF & kNN & ALR  & SVR & synth. \\
\hline\hline
Gaze error (M)\degree & \textbf{8.85} & 9.44 & 10.5 & 12.0 & 12.2 & 12.7 & 15.1 & 19.9 \\
\hline
\end{tabular}
\vspace{0.2cm}
\caption{Our model outperforms eye3DMM \cite{Wood2016A3M}, CNN \cite{7299081}, Random Forests (RF) \cite{Sugano:2014:LAG:2679600.2680301}, kNN \cite{7299081}, Adaptive Linear Regression (ALR) \cite{6777326}, and Support Vector Regression (SVR) \cite{6976920} in mean gaze estimation error on the Eyediap database \cite{FunesMora_ETRA_2014}.}
\label{tab:gaze_stimation}
\end{table*}


\section{Conclusion}
\label{sec:conclusions}
In this work, we propose the first human head 3DMM representation that is complete in the sense that it demonstrates meaningful variations across all major visible surfaces of the human head - that is face, cranium, ears and eyes. In addition, for realistic renderings in open-mouth expressions, a basic model of the oral cavity, tongue and teeth is included. We presented a pipeline to fuse multiple 3DMMs into a single 3DMM and used it to combine the LSFM face model, the LYHM head model, and a high-detail ear model. Furthermore, we incorporated a detailed eye model that is capable of reconstructing accurately the eyelid shape and the shape around the eyes as well as the eye gaze and color. Additionally, we build a complete high-detail head texture model by constructing a framework that enables us to complete the missing head texture for any given facial texture. The resulting universal head model captures all the desirable properties of the constituent 3DMMs; namely, the high facial detail of the facial model, the full cranial shape variations of the head model, the additional high quality ear variations as well as the bespoke eyelid and eye region deformations. The augmented model is capable of representing and reconstructing any given head shape (including ears  and eyelid shape) due to the high variation of facial and head appearances existing in the original models. We demonstrated that our methodology yielded a statistical model that is considerably superior to the original constituent models. Finally we illustrated the model's utility in full head reconstruction from a single images.

Although our model is a significant step forward, the challenge of a universal head model remains open. We do not deal with hair, instead modelling cranium geometry with skin texture and baking facial hair into the texture. We do not fully model the statistical shape variance inside the mouth, including teeth and tongue, which is essential for realistic speech dynamics. Rather, we only statistically model external craniofacial shape. There may be value in modelling internal skull geometry and a volumetric skin model both for disentangling rigid body motion from face dynamics and also to enable more accurate rendering. Finally, we still depend on a classical shape modelling pipeline of GPA and PCA where more sophisticated, nonlinear models may be preferable.

\ifCLASSOPTIONcompsoc
  \section*{Acknowledgments}
\else
  \section*{Acknowledgment}
\fi

S.~Ploumpis was suppored by EPSRC Project \small{(EP/N007743/1)} FACER2VM. S.~Zafeiriou acknowledges funding from a Google Faculty Award, as well as from the EPSRC Fellowship DEFORM: Large Scale Shape Analysis of Deformable Models of Humans \small{(EP/S010203/1)}. N.~Pears and W.~Smith acknowledge funding from a Google Daydream award. W.~Smith was supported by the Royal Academy of Engineering under the Leverhulme Trust Senior Fellowship scheme. We would like to formally thank Vasileios Triantafyllou (computer graphics specialist at Facesoft) for all the visual content of this work, as well as for his valuable contribution to the head texture completion dataset.  

\ifCLASSOPTIONcaptionsoff
  \newpage
\fi


\bibliographystyle{IEEEtran}
\bibliography{egbib}

\input{biographies.tex}

\end{document}

%% file: biographies.tex








\begin{IEEEbiography}[{\includegraphics[width=1in,height=1.25in,clip,keepaspectratio]{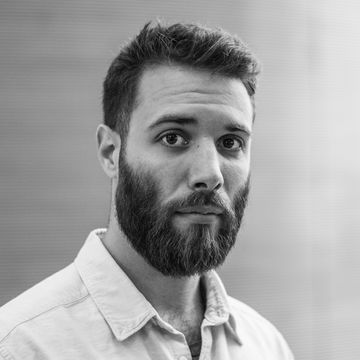}}]{Stylianos Ploumpis}
received the Diploma and Master of Engineering in Production Engineering \& Management from Democritus University of Thrace, Greece (D.U.T.H.), in 2013. He joined the department of computing at Imperial College London, in Octomber 2015, where he pursued an MSc in Computing specialising in Machine Learning. Currently, he is a PhD candidate/Researcher at the Department of Computing at Imperial College, under the supervision of Dr. Stefanos Zafeiriou. His research interests lie in the field of 3D Computer Vision, Pattern Recognition and Machine Learning.
\end{IEEEbiography}

\begin{IEEEbiography}[{\includegraphics[width=1in,height=1.25in,clip,keepaspectratio]{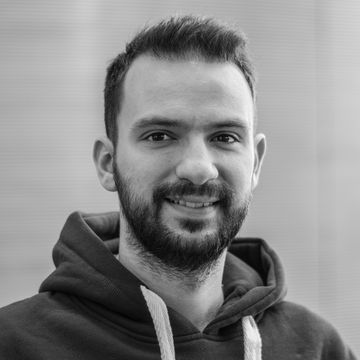}}]{Evangelos Ververas}
graduated in September 2016 from the Department of Electrical and Computer Engineering in Aristotle University of Thessaloniki, in Greece. He joined the Department of Computing at Imperial College London in October 2016 and he is currently working as a PhD Student/Teaching Assistant under the supervision of Dr. Stefanos Zafeiriou. His research focuses on 3D reconstruction.
\end{IEEEbiography}

\begin{IEEEbiography}[{\includegraphics[width=1in,height=1.25in,clip,keepaspectratio]{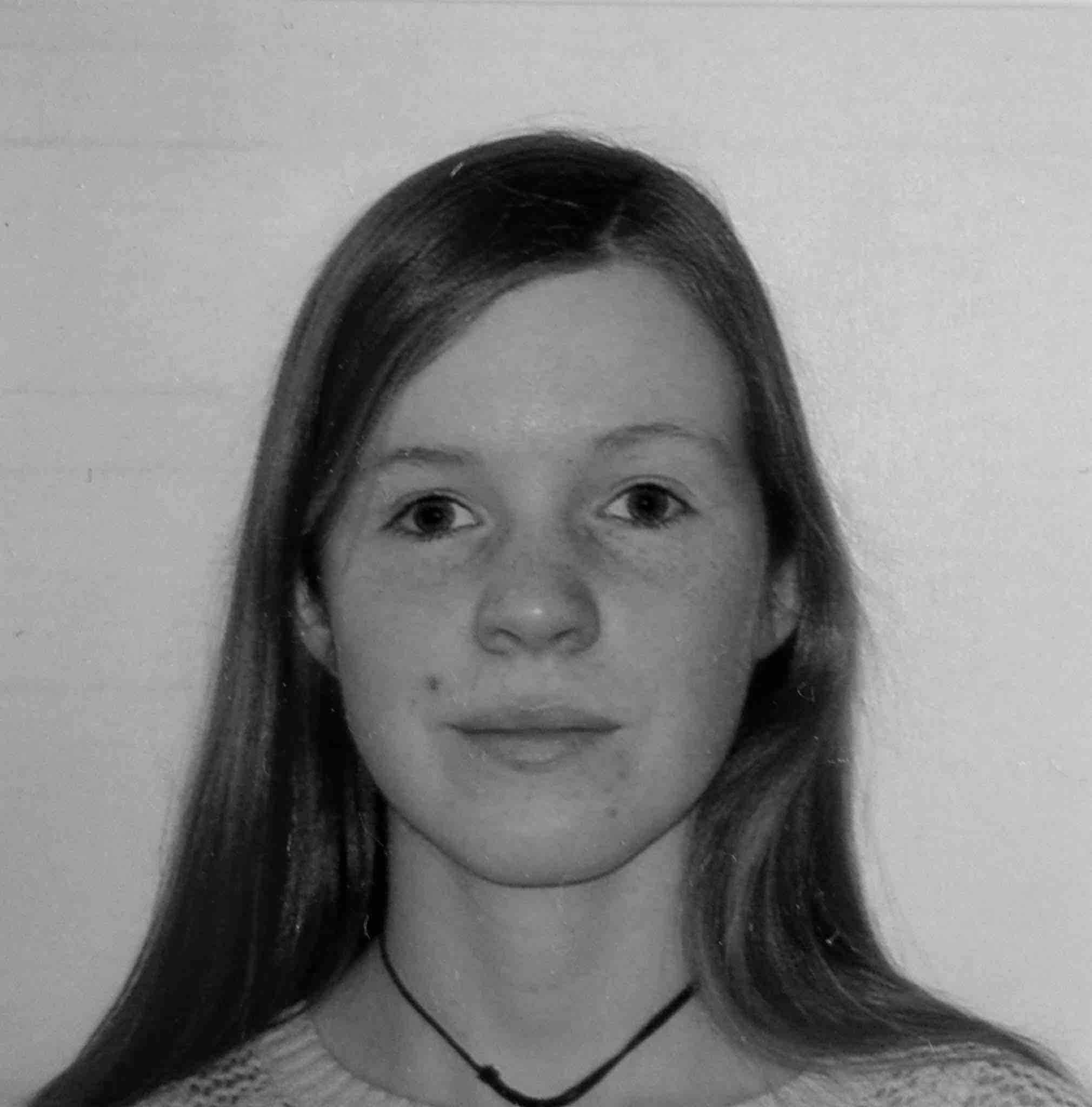}}]{Eimear O’ Sullivan}
received the BEng in Electronic and Computer Engineering from the University of Limerick, Ireland, and the MRes in Medical Robotics and Image Guided Intervention from Imperial College London in 2016 and 2018 respectively. She joined the Department of Computing, Imperial College London in October 2018 and is working towards the PhD degree under the supervision of Dr. Stefanos Zafeiriou. Her research interests include geometric deep learning and computer vision for the analysis and characterization of craniosynostosis.
\end{IEEEbiography}

\begin{IEEEbiography}[{\includegraphics[width=1in,height=1.25in,clip,keepaspectratio]{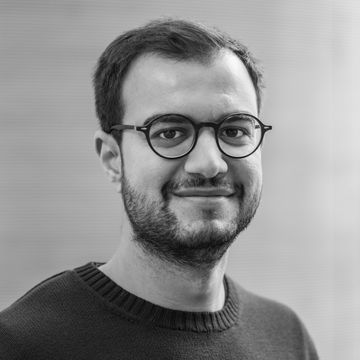}}]{Stylianos Moschoglou}
 received his Diploma/MEng in Electrical and Computer Engineering from Aristotle University of Thessaloniki, Greece, in 2014. In 2015-16, he pursued an MSc in Computing (specialisation Artificial Intelligence) at Imperial College London, U.K., where he completed his project under the supervision of Dr. Stefanos Zafeiriou.

He is currently a PhD student at the Department of Computing, Imperial College London, under the supervision of Dr. Stefanos Zafeiriou. His interests lie within the area of Machine Learning and in particular in Generative Adversarial Networks and Component Analysis.
\end{IEEEbiography}

\begin{IEEEbiography}[{\includegraphics[width=1in,height=1.25in,clip,keepaspectratio]{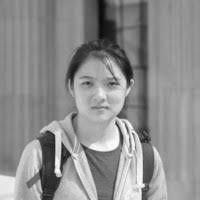}}]{Haoyang Wang}
 received a bachelor degree on Computer Science with Artificial Intelligence from the University of Nottingham, and a MSc degree on Machine Learning from Imperial College London. She is currently a PhD student at the Department of Computing, Imperial College London, under the supervision of Dr. Stefanos Zafeiriou. Her research interest is 3D computer vision for human face, body reconstruction, shape and pose estimation.
\end{IEEEbiography}

\begin{IEEEbiography}[{\includegraphics[width=1in,height=1.25in,clip,keepaspectratio]{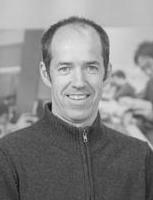}}]{Nick Pears}
is an Associate Professor in Computer Science at the University of York, UK. He works in Computer Vision and Machine Learning - particularly medical, biometric and creative applications that employ statistical 3D shape modelling. He was recently awarded a Senior Research Fellowship award by the Royal Academy of Engineering (UK) and has had his recent work supported by two Google Faculty Awards. Previously he has worked at the University of Cambridge and University of Oxford, UK, and he gained his BSc in Engineering Science and PhD in Robotics from the University of Durham, UK.
\end{IEEEbiography}

\begin{IEEEbiography}[{\includegraphics[width=1in,height=1.25in,clip,keepaspectratio]{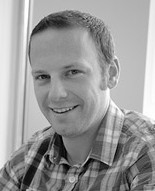}}]{William A. P. Smith}
received the BSc degree in computer science, and the PhD degree 
in computer vision from the University of York, York, United Kingdom. He is currently a Reader with the Department of Computer Science, University of York, York, United Kingdom and a Royal Academy of Engineering/The Leverhulme Trust Senior Research Fellow. His research interests are in shape and appearance modelling, model-based supervision and physics-based vision. He has published more than 100 papers in international conferences and journals.
\end{IEEEbiography}

\begin{IEEEbiography}[{\includegraphics[width=1in,height=1.25in,clip,keepaspectratio]{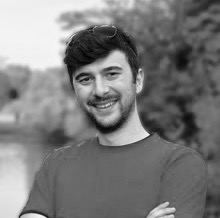}}]{Baris Gecer} is a PhD. student in the Department of Computing, Imperial College London, under the supervision of Dr. Stefanos Zafeiriou. His main research interests are photorealistic 3D Face modelling and synthesis by Generative Adversarial Nets and Deep Learning. He obtained his M.S. degree from Bilkent University Computer Engineering department under the supervision of Prof. Selim Aksoy in 2016 and obtained his undergraduate degree in Computer Engineering from Hacettepe University in 2014.
\end{IEEEbiography}

\begin{IEEEbiography}[{\includegraphics[width=1in,height=1.25in,clip,keepaspectratio]{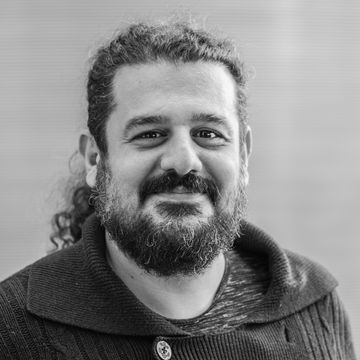}}]{Stefanos Zafeiriou}
is a Reader in Machine Learning for Computer Vision with the Department of Computing, Imperial College London, and a Distinguishing Research Fellow with University of Oulu under Finish Distinguishing Professor Programme.
He was the recipient of a prestigious Junior Research Fellowship from Imperial College London in 2011 to start his own independent research group.
\end{IEEEbiography}

\vfill